\documentclass[10pt,journal,compsoc]{IEEEtran}

\newcommand{\his}{\mbox{$\mathbf{g}$}\xspace}

\newcommand{\PGateBeta}{\mbox{$\mathbf{q^\top}$}\xspace}
\newcommand{\RGateBeta}{\mbox{$\mathbf{c^\top}$}\xspace}

\newcommand{\Sets}{\mbox{$B$}\xspace}
\newcommand{\Set}{\mbox{$b$}\xspace}

\newcommand{\method}{\mbox{$\mathop{\mathtt{M^{2}\text{-}gp^2t}}\limits$}\xspace}
\newcommand{\methodAll}{\mbox{$\mathop{\mathtt{M^2}}\limits$}\xspace}
\newcommand{\transModel}{\mbox{$\mathop{\mathtt{ed\text{-}Trans}}\limits$}\xspace}

\newcommand{\POP}{\mbox{$\mathop{\mathtt{POP}}\limits$}\xspace}
\newcommand{\POEP}{\mbox{$\mathop{\mathtt{POEP}}\limits$}\xspace}
\newcommand{\FREQ}{\mbox{$\mathop{\mathtt{M^{2}\text{-}p^2}}\limits$}\xspace}
\newcommand{\FREQP}{\mbox{$\mathop{\mathtt{M^{2}\text{-}gp^2}}\limits$}\xspace}
\newcommand{\Dream}{\mbox{$\mathop{\mathtt{Dream}}\limits$}\xspace}
\newcommand{\FPMC}{\mbox{$\mathop{\mathtt{FPMC}}\limits$}\xspace}
\newcommand{\SetsSets}{\mbox{$\mathop{\mathtt{Sets2Sets}}\limits$}\xspace}

\newcommand{\GP}{\mbox{$\mathop{\mathtt{UGP}}\limits$}\xspace}
\newcommand{\IGP}{\mbox{$\mathop{\mathtt{IGP}}\limits$}\xspace}
\newcommand{\IT}{\mbox{$\mathop{\mathtt{TPI}}\limits$}\xspace}
\newcommand{\IC}{\mbox{$\mathop{\mathtt{CPI}}\limits$}\xspace}

\newcommand{\xia}[1]{\textcolor{red}{Xia comments: #1}}

\usepackage{enumitem}
\usepackage{xr}
\usepackage[normalem]{ulem}
\usepackage{subcaption}
\usepackage{graphicx}
\usepackage{comment}
\usepackage{amsmath}
\usepackage{amsfonts}
\usepackage{amssymb}
\usepackage{algorithm}
\usepackage{algpseudocode}
\usepackage{pgfplotstable}
\usepackage{booktabs} 
\usepackage{longtable}
\usepackage{tikz}
\usetikzlibrary[automata, positioning]
\usepackage{hyperref}
\hypersetup{colorlinks=false}
\usepackage[flushleft]{threeparttable}
\usepackage{multirow}
\usepackage{bbm}
\usepackage{caption}
\usepackage{xspace}
\usepackage{verbatim,soul}
\usepackage{rotating}
\usepackage{fixltx2e}
\usepackage{xurl}
\usepackage{mathtools}
\usepackage{color, colortbl}
\usepackage{amsmath}
\usepackage{footmisc}
\newcommand{\etal}{\textit{et al}.}
\definecolor{aureolin}{rgb}{0.99, 0.93, 0.0}
\DeclareMathOperator*{\argmin}{argmin}

\ifCLASSOPTIONcompsoc
  \usepackage[nocompress]{cite}
\else
  \usepackage{cite}
\fi

\begin{document}

\title{\methodAll: Mixed Models with Preferences, Popularities and Transitions for Next-Basket Recommendation}


\author{Bo~Peng,
        Zhiyun~Ren,
        Srinivasan~Parthasarathy,~\IEEEmembership{Member,~IEEE,}
        and~Xia~Ning$^*$,~\IEEEmembership{Member,~IEEE}
\IEEEcompsocitemizethanks{
\IEEEcompsocthanksitem Bo Peng 
is with the Department
of Computer Science and Engineering, The Ohio State University, Columbus,
OH, 43210.\protect\\
E-mail: peng.707@buckeyemail.osu.edu 
\IEEEcompsocthanksitem Srinivasan Parthasarathy and Xia Ning are with the Department of Biomedical Informatics, the Department of Computer Science and Engineering, 
and the Translational Data Analytics Institute,
The Ohio State University, Columbus, OH, 43210.\protect\\
E-mail:  srini@cse.ohio-state.edu, ning.104@osu.edu
\IEEEcompsocthanksitem Zhiyun Ren is with the Department of Biomedical Informatics, 
The Ohio State University, Columbus, OH, 43210.\protect\\
E-mail: ren.685@osu.edu
\IEEEcompsocthanksitem $^*$Corresponding author
}
\thanks{Manuscript received April 19, 2005; revised August 26, 2015.}}

\markboth{Journal of \LaTeX\ Class Files,~Vol.~14, No.~8, August~2015}%
{Shell \MakeLowercase{\textit{et al.}}: Bare Demo of IEEEtran.cls for Computer Society Journals}
%

%

\IEEEtitleabstractindextext{%
\begin{abstract}
Next-basket recommendation considers the problem of recommending a set of items 
into the next basket that users will purchase as a whole.
In this paper, 
we develop a novel mixed model with preferences, popularities and transitions (\methodAll) for the next-basket recommendation.
This method models three important factors in next-basket generation process: 
1) users' general preferences, 2) items' global popularities and 3) transition patterns among items. 
Unlike existing recurrent neural network-based approaches, {\methodAll} does not use the complicated networks to 
model the transitions among items, or generate embeddings for users.
Instead, it has a simple encoder-decoder based approach ({\transModel}) to better model the transition patterns among items.
We compared \methodAll with different combinations of the factors with 5 state-of-the-art
next-basket recommendation methods on 4 public benchmark datasets 
in recommending the first, second and third next basket. 
Our experimental results demonstrate that {\methodAll} significantly outperforms the state-of-the-art methods 
on all the datasets in all the tasks, with an improvement of up to 22.1\%.
In addition, our ablation study demonstrates that the ed-Trans is more effective than recurrent neural networks in terms of the recommendation performance.
We also have a thorough discussion on various experimental protocols and evaluation metrics
for the evaluation of next-basket recommendation methods. 


%

\end{abstract}

}
\maketitle

\IEEEdisplaynontitleabstractindextext
\IEEEpeerreviewmaketitle

\IEEEraisesectionheading{\section{Introduction}}
\label{sec:intro}

\IEEEPARstart{N}{ext-basket} recommendation~\cite{sets2sets,fpmc,dream,bai2018attribute,yang2019pre} considers the problem of recommending a set of items 
into the next basket that users will purchase as a whole, based on the baskets of items that users have purchased. 
It is different from the conventional top-$N$ recommendation 
problem in recommender systems, in which users will purchase a single item at each time. Next-basket recommendation has been drawing increasing 
attention from research community due to its wide applications in the grocery industry~\cite{sets2sets,dream}, fashion industry~\cite{fossil} and 
tourism industry~\cite{beacon}, etc.
%
%
With the prosperity of deep learning, many deep models, particularly based on recurrent neural networks (RNNs)~{\mbox{\cite{sets2sets,dream,bai2018attribute,yang2019pre}}} 
have been developed for next-basket recommendation purposes, and have demonstrated superior performance~\mbox{\cite{sets2sets,dream}}.
These methods, especially these RNN-based methods, often focus on modeling the transitions between different baskets, 
but are not always effective to model various important factors that may determine 
next baskets.
For example, the transition among individual items in different baskets
is an important factor, 
as given the individual items in the previous baskets, the probability of being interacted/purchased 
in the next basket is not equal for all the items.
Users' general preference is another important factor as different users generally
will have different preferences on items.
%
Recently developed RNN-based methods~\cite{sets2sets,dream,bai2018attribute} typically explicitly model the transitions among baskets, 
while implicitly model the transitions among individual items.
For example, these methods use mean pooling or weighted sum to aggregate the items in a same basket, 
and then use the recurrent units to model the transitions among baskets.
However, during such aggregation, the information of individual items could be smoothed out 
so that these methods could not accurately model the transitions among individual items.
In addition, due to the recurrent nature of RNNs, it is challenging to train these 
RNN-based methods efficiently in parallel. 
%
Another limitation with existing methods is that
learned user embeddings are usually employed as the representations of users' general preferences. 
However, due to the notoriously sparse nature of data in recommendation problems, 
these learned embeddings may not be able to accurately capture users' preferences.
%
To mitigate the limitations in the existing basket recommendation methods,
in this paper, we develop a set of novel mixed models, denoted as \methodAll, 
for the next-basket recommendation problem.

\methodAll 
models three
important factors in order to generate next-basket recommendations for each user.
The first factor is users' general preferences, which will measure long-term preferences of users 
that tend to remain consistent 
across multiple baskets during a certain period of time.
The second factor is the global popularities of items, which will measure the overall
popularities of items among all the users.
The third factor is the transition patterns among items across baskets, which will capture the 
transition patterns on items over different baskets.
%
%
These three factors will be combined together using weights that will be determined by 
these factors, 
and thus recommend items into the next basket. 
With different combinations of factors, \methodAll has three variants \FREQ, \FREQP and \method.
\FREQ recommends items using users' general preferences and items' global popularities. 
In \FREQ, these two factors are combined using a global weight.
\FREQP is similar to \FREQ 
except that instead of using a global weight, {\FREQP} learns personalized weights to combine the two factors.
\method uses all the three factors for more accurate recommendations.
%
The details of these three variants will be presented in Section~\ref{sec:method}.
In particular, different from existing methods, 
\methodAll explicitly models the transitions among individual items 
using a simple{, efficient, and effective}
encoder-decoder based framework, denoted as \transModel.
{\methodAll} also explicitly models users' general preferences using 
the frequencies of items that each user has interacted with instead of the user embeddings.

%
We compare \methodAll with 5 most recent, state-of-the-art methods on 4 public benchmark datasets 
in recommending the first, second and third next basket.
Our experimental results demonstrate that \methodAll significantly outperforms the state-of-the-art methods 
on all the datasets in all the tasks, 
with an improvement of up to 22.1\%.
%
We also conduct a comprehensive ablation study to verify the effects of the different factors.
The results of the ablation study show that 
learning all the factors
together could significantly improve the recommendation performance compared to learning
each of them alone.
The results also show that the encoder-decoder based \transModel  
in learning item transitions among baskets
could outperform RNN-based methods on the benchmark datasets. 

%
The major contributions in this paper are as follows:
\begin{itemize}[noitemsep,nolistsep,leftmargin=*]
\item We developed a novel mixed model \methodAll   
for next-basket recommendation.
\methodAll explicitly models three important factors: 1) users' general preferences, 
2) items' global popularities, and 
3) transition patterns among items. 
\item We developed a novel, simple yet effective encoder-decoder based framework \transModel to 
model transition patterns among items in baskets. 
\item \methodAll significantly outperforms state-of-the-art methods. Our experimental results 
over 4 benchmark datasets demonstrate that \methodAll achieves significant improvement in 
both recommending the next basket and recommending the next a few baskets, with an improvement 
as much as 22.1\%. 
Our ablation study shows that the factors are complementary 
and enable better performance if learned together (Section~\ref{sec:results:ablation}).
\item Our ablation study also shows that \transModel
in learning item transitions among baskets
could on its own significantly outperform RNN-based methods over the benchmark datasets, 
with an improvement as much as 25.4\% (Section~\ref{sec:results:ablation:rnn}). 
\item Our cluster analysis shows that \transModel is able to learn similar embeddings 
for items which have similar transition patterns (Section~\ref{sec:results:clusters}).
\item We discussed the potential issues of evaluation metrics, experimental protocols 
and settings that are typically 
used in next-basket recommendation, and discussed the use of 
a more appropriate protocol and setting in our experiments (Section~\ref{sec:discusion}).   
\item For reproducibility purposes, we released our source code and Supplementary Materials at \url{https://github.com/BoPeng112/M2}. 
\end{itemize}

\section{Related Work}
\label{sec:literature}

\subsection{Next-Basket Recommendation}
\label{sec:literature:basket}
Numerous next-basket recommendation methods have been developed,
particularly using 
Markov Chains (MCs) and Recurrent Neural Networks (RNNs) 
etc. 
Specifically, MCs-based methods, such as factorized personalized Markov chains (\FPMC)~\cite{fpmc}, 
use MCs to model the pairwise item-item transition patterns 
to recommend the next item or the next basket of items for each user.
%
Wan \etal~\cite{adaloyal} developed factorization-based methods \mbox{triple2vec} 
and \mbox{adaLoyal}, 
in which the item-item complementarity, user-item compatibility and user-item loyalty patterns are 
modeled for the next-basket recommendation. 
Recently, RNN-based methods have been developed for the next-basket recommendation.
For instance, Yu \etal~\cite{dream} used RNNs to model users' dynamic short-term preference at different timestamps.
Wang \mbox{\etal}~\mbox{\cite{Wang2021}} developed a hierarchical attentive encoder-decoder model, 
which iteratively predicts the next baskets by learning the transitions among items and leveraging both the positive and negative feedbacks 
from users.
Hu \etal~\cite{sets2sets} developed an encoder-decoder RNN method \SetsSets. 
\SetsSets employs an RNN as encoder to learn users' dynamic preference at different timestamps 
and another RNN as decoder to generate the recommendation score from the learned preferences 
for each recommendation candidate. 
\SetsSets has been demonstrated as the state of the art, and outperforms an extensive set of existing methods. 

%
Aside from model-based methods, 
popularity-based approaches such as popularity 
on people ({\POP})~\mbox{\cite{sets2sets}} and popularity on each person ({\POEP})~\mbox{\cite{sets2sets}}, are also recently employed
for the next-basket recommendation.
\POP ranks items based on their popularity among all the users 
and recommend the top-$k$ most popular items to each user.
\POEP is the personalized version of \POP. 
It ranks items based on their popularity of each user and recommends the top-$k$ most popular items of each user.
These two popularity-based methods have been demonstrated as strong baselines
 on the next-basket recommendation in the recent work~\cite{sets2sets}.
 %
 
 Unlike existing RNN-based approaches,
 {\methodAll} does not use the Markov chains or complicated RNNs to model the transitions among items, or
 generate embeddings for users. 
 Instead, {\methodAll} models the transitions among items using a simple yet effective fully-connected layer, and explicitly 
 models users' general preferences as the frequencies of items that users have interactions with.
 Our experimental results demonstrate the superior performance 
 of {\methodAll} over the state-of-the-art baseline methods. 
 Our ablation study also shows that the fully-connected layer is more effective 
 than RNNs in terms of the recommendation performance.

\subsection{Sequential Recommendation}
\label{sec:literature:sequential}

Sequential recommendation is to generate the recommendation of the next items 
based on users' historical interactions as in a sequence.
This task is closely related to the next-basket recommendation.
Please refer to 
the Supplementary 
Materials~\footnote{Section references starting with ``S" refer to the sections in the Supplementary Materials.} 
for a detailed discussion about the relations between 
these two tasks.
The sequential recommendation methods focus on capturing the sequential dependencies among 
individual items instead of baskets.
In the last few years, numerous sequential recommendation methods have been developed, 
particularly using neural networks such as Recurrent Neural Networks (RNNs), Convolutional Neural Networks (CNNs) 
and attention or gating mechanisms, etc.
RNN-based methods such as User-based RNN~\mbox{\cite{ugru}} explicitly integrates user characteristics into gated recurrent units (GRUs) 
for personalized recommendation.
Skip-gram-based methods such as item2vec~\cite{itemvec} 
and prod2vec~\cite{prodvec} leverage the skip-gram model~\cite{wordvec} to learn 
transition patterns among individual items.
Recently, CNN-based and attention-based methods have been developed for sequential recommendation.
For example, Tang \etal~\cite{caser} developed a convolutional sequence embedding recommendation model (Caser), 
which uses convolutional filters on the most recent items to extract union-level features. 
Kang \etal~\cite{sasrec} developed a self-attention based sequential model (SASRec), 
which uses attention mechanisms to capture the most informative items in users' historical interactions 
to generate recommendations. 
Sun \mbox{\etal}~\mbox{\cite{bert4rec}} further developed a bidirectional self-attention based sequential 
model (BERT4Rec), which employs a bidirectional attention mechanism to better model users' historical interactions.
Recently, Ma \etal~\cite{hgn} developed a hierarchical gating network (HGN), which uses gating mechanisms to identify important items and generate recommendations.
Peng \etal~\cite{ham} developed hybrid associations models (HAM), 
which adapt the pooling mechanisms to model the association patterns and synergies among items.

\subsection{Session-based Recommendation}
\label{sec:literature:session}

Session-based recommendation seeks to generate the recommendations of the next items in the current session or future sessions 
based on users' interactions in historical sessions. 
This task is also closely related to the next-basket recommendation.
The session-based recommendation methods focus on capturing the intra- or inter-session dependencies to generate the recommendations~\mbox{\cite{session}}.
In the last few years, neural networks such as RNNs, attention mechanisms and graph neural networks (GNNs) 
are employed in developing session-based recommendation methods.
RNN-based methods such as GRU4Rec~\mbox{\cite{gru4rec}} and GRU4Rec+~\mbox{\cite{gru4rec+}} 
employ gated recurrent units (GRUs) to capture the users' dynamic short-term preferences over sessions.
%
Attention-based methods such as NARM~\mbox{\cite{narm}} and STAMP~\mbox{\cite{stamp}} employ attention mechanisms to identify the important items in 
recent sessions to capture users' short-term preferences.
Recently, GNN-based methods have also been developed for the session-based recommendation.
For example, Wu \mbox{\etal}~\mbox{\cite{srgnn}} developed a GNN-based recommendation model (SR-GNN) to 
better model the long-term dependency among sessions.
Qiu \mbox{\etal}~\mbox{\cite{fgnn}} re-examined the item ordering in session-based recommendations 
and developed a GNN-based model (FGNN) to identify the items representing 
users' short-term preferences in sessions.

\section{Definitions and Notations}
\label{sec:definitions}

\begin{table}[!t]
  \caption{Notations}
  \label{tbl:notations}
  \vspace{-5pt}
  \centering
  \begin{threeparttable}
      \begin{tabular}{
	@{\hspace{2pt}}l@{\hspace{2pt}}
	@{\hspace{2pt}}l@{\hspace{2pt}}          
	}
        \toprule
        notations & meanings \\
        \midrule
        $m$/$n$ &  number of users/items \\
        $d$ & dimension of latent representation of next basket \\
        $\Sets_i$/$\Set_i(t)$ & the basket sequence/the $t$-th basket of user $i$ \\
        $T_i$/$n_i(t)$ & the number of baskets in $\Sets_i$/of items in $\Set_i(t)$\\
       $\mathbf{r}_i$ & the vector representation of the basket $\Set_i(T_i)$\\
        $\hat{\mathbf{r}}_{i}$ & the recommendation scores over all items for user $i$\\
        \bottomrule
      \end{tabular}
  \end{threeparttable}
\end{table}


%
In this paper, the historical interactions (e.g., purchases, check-ins) of the $i$-th user in chronological order 
are represented as a sequence of baskets $\Sets_i$$=$$\{\Set_i(1),\Set_i(2), \cdots\}$, 
where $\Set_i(t)$ is a basket of one or more items in the $t$-th interaction.
Note that there may be multiple, same items in each basket (e.g., three apples in one basket).
The number of baskets in $\Sets_i$ and the number of items in $\Set_i(t)$ 
is denoted as $T_i$ and $n_i(t)$, respectively.
In this paper, we consider all the baskets in users' history and all the items in each basket.
We do not have a predefined maximum length for the basket sequences, 
and maximum size for each basket. 
When no ambiguity arises, we will eliminate $i$ in $\Sets_i$/$\Set_i(t)$, $T_i$ and $n_i(t)$.
In this paper, all the vectors are by default row vectors and represented using lower-case bold letters;
all the matrices are represented using upper-case letters.
The key notations are 
in Table~\ref{tbl:notations}.

\section{Methods}
\label{sec:method}

\subsection{Modeling Important Factors in \methodAll}
\label{sec:method:factors}


%
\begin{figure}[!t]
    \centering
   \includegraphics[width=1.03\linewidth]{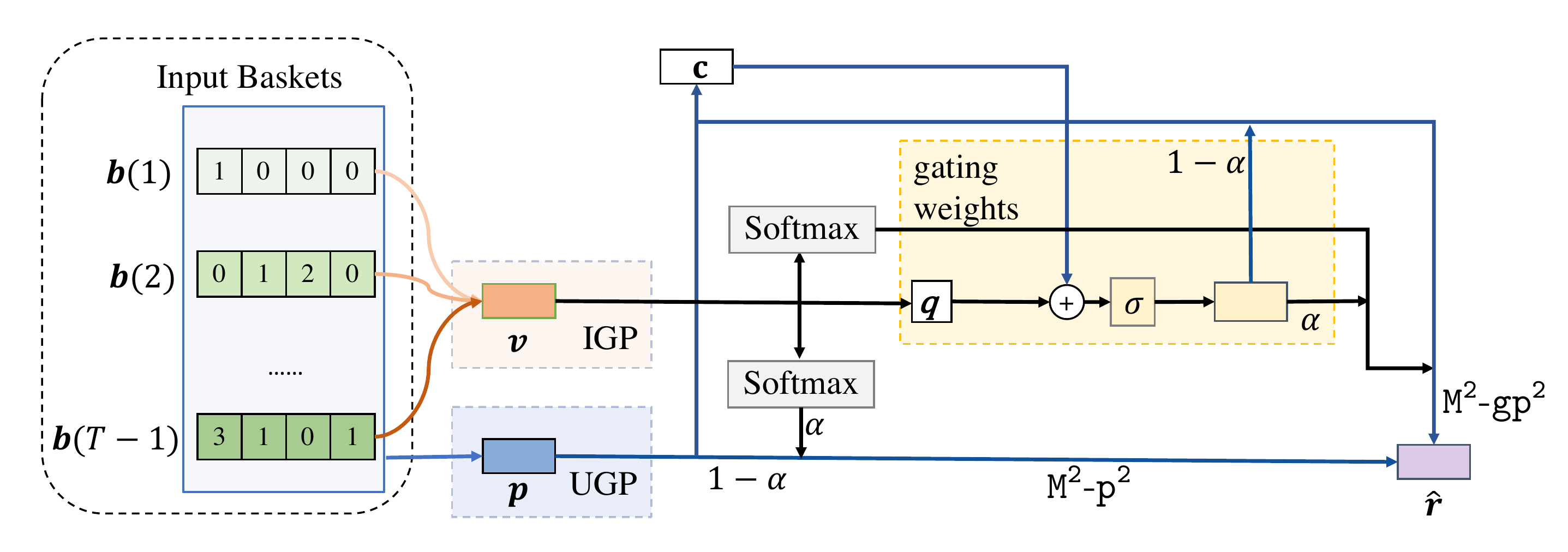}
    \caption{\FREQ and \FREQP Model Architectures}
    \label{fig:architectureFREQ}
%
    \vspace{10pt}
%
    \centering
   \includegraphics[width=0.95\linewidth]{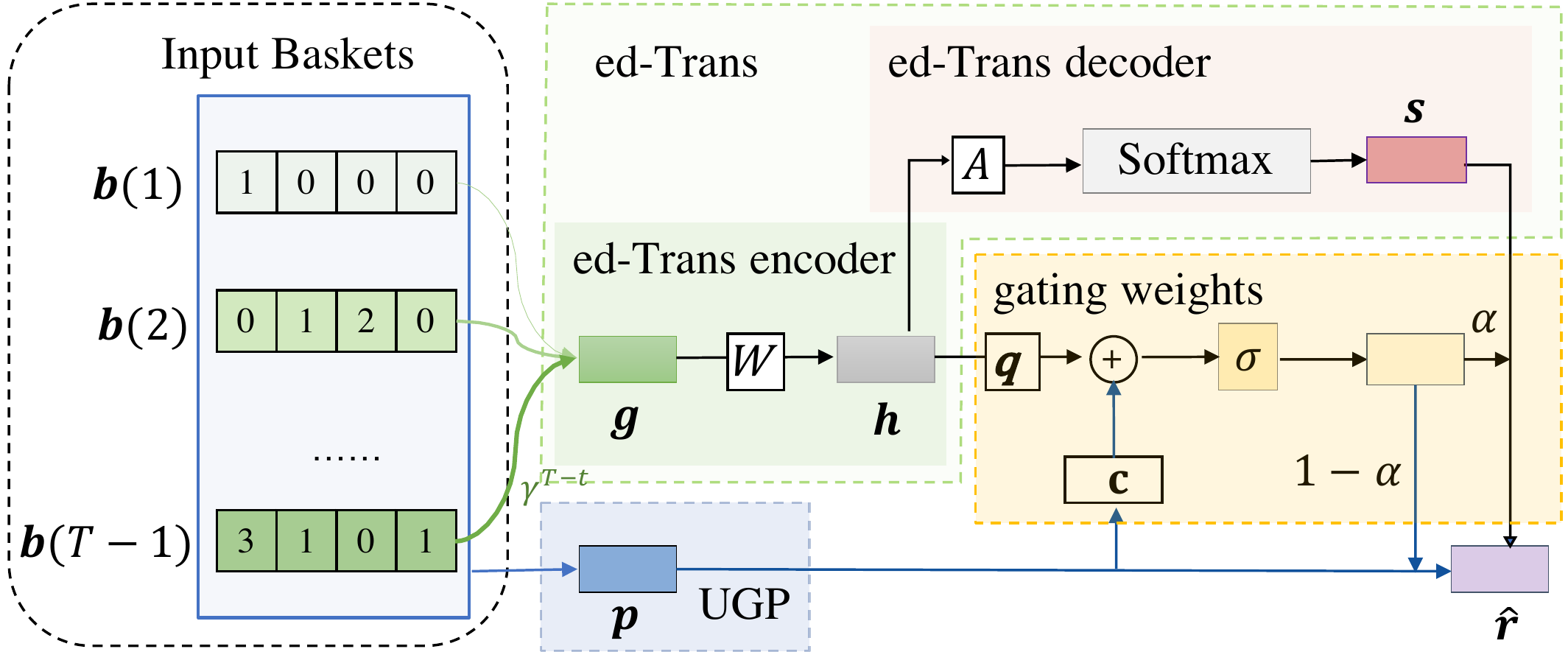}
    \caption{\method Model Architecture}
    \label{fig:architecture}
\end{figure}

\methodAll has three variants \FREQ, \FREQP and \method.
Figure~\ref{fig:architectureFREQ} presents the  \FREQ and \FREQP models.
Figure~\ref{fig:architecture} presents the \method model. 
In these figures, each input basket is represented as a vector of  $n$ (i.e., the number of items) dimensions, 
in which the value in each dimension represents the number of the corresponding item in this basket.
\methodAll generates recommendations
\noindent
for the next baskets of items 
for each user using three factors: 
1) users' general preferences, 2) items' global popularities
and 3) the transition patterns among items across baskets.
These three factors will be used to calculate a recommendation score for 
each candidate item in the next baskets. In this section, we will first describe how these three factors 
are modeled. In the next Section, we will describe how the three variant methods use these factors for recommendations. 

%

\subsubsection{Modeling Users' General Preferences (\GP)}
\label{sec:method:factors:general}
Previous studies have shown that users' interactions are significantly affected
by their general preferences~\cite{fossil,sets2sets}, 
which are also known as the long-term preferences in the literature~\mbox{\cite{ham,fossil}}. 
For example, some users prefer items of low price, while others may like luxurious items 
that could be expensive.
Therefore, we explicitly model the general preferences of users, denoted as \GP, 
in \methodAll.
Existing methods~\cite{fpmc} usually model users’ general preferences using the embeddings of users.
However, there is limited, if any, validation showing that the learned embeddings could 
accurately capture users’ preferences and to what extent.
%
Thus, in \methodAll, 
we propose to use the frequencies of items that each user has interactions with 
to represent users' general preferences.
%
The intuition is that if a user has many interactions with an item, 
the user has a high preference on the item and the item represents the user's preference. 

Specifically, given each user's historical interactions, her/his general preference is represented as follows,
\begin{equation}
\label{eqn:p}
\mathbf{p} = [p_1, p_2, \cdots, p_n]\in \mathbb{R}^{1 \times n}, 
\end{equation}
where 
%
\begin{equation}
\label{eqn:general}
{p}_{j} = {n_{j}}/{\sum\nolimits_j n_{j}}, 
\end{equation}
%
$n$ is the total number of unique items among all the baskets, 
and $n_{j}$ is the total number of interactions with item $j$ of the user among all her/his interactions,
and thus $p_{j} \ge 0$, $\sum_j p_{j} = 1$. 
Here, we do not weight the interactions on items 
differently based on when they occur. This is because
in real applications, we typically only use the data in the 
past, relatively short period of time (e.g., a few months) to train recommendation 
models{~\mbox{\cite{pinner}}}. 
In this short period, we can assume that most of the users will not change their general 
preferences dramatically, and thus all their interacted items will contribute to their 
\GP estimation evenly. 
A distinct advantage of the preference representation \GP as in Equation~\ref{eqn:p} compared to 
embedding representations for user preferences is that the \GP representation is very intuitive and easy to validate, 
and loses minimum user information. 

The formulation of users' general preferences in Equation~{\ref{eqn:p}}
is designed for the application scenarios that users are likely to have
multiple interactions with the same item (e.g., online shopping, grocery shopping).
For the other few application scenarios that do not have this property
 (e.g., movie recommendation), our formulation may not be applicable.
We leave the investigation of these applications in the future work.

\subsubsection{Modeling Items' Global Popularities  (\IGP)}
\label{sec:method:factors:popularity}

It has been shown in the literature~\cite{fpmc,sets2sets,mcfm,sarwar2001item} that the items' global popularities also 
significantly influence users' purchases. 
Specifically, users may prefer popular items than those non-popular ones due to the herd behaviors~\cite{herd}, 
that is, they prefer to purchase items that are also purchased by many others.
%
In \methodAll, 
the items' global popularities are represented as in the following vector $\mathbf{v}$, 
\begin{equation}
\label{eqn:v}
\mathbf{v} = [v_1, v_2, \cdots, v_n]\in \mathbb{R}^{1 \times n}, 
\end{equation}
where $n$ is the total number of unique items among all the baskets, and $v_j$ is a learnable scalar 
to represent the global popularities of item $j$. 
Intuitively, if item $j$ is popular, $v_j$ will be large.
Here, following the ideas in Koren \mbox{\etal}~\mbox{\cite{mcfm}}, we learn the popularity representations (i.e., $v_j$) for 
items via learning and optimizing from data for better performance rather than directly calculating them from data.

\subsubsection{Modeling Transitions among Items (\IT) via an Encoder-Decoder Framework (\transModel)} 
\label{sec:method:itemsTrans}

The transitions among items is another important factor in inducing the next baskets of 
items that the users will be interested in~\cite{fpmc,caser,ham}.
For example, if a user purchased cat toys in a basket, she/he is likely to purchase cat food and treats
in the next baskets compared to wine and beers, as there could be stronger transitions among cat items compared 
to from cat items to alcohols.
%
%
In \methodAll, we explicitly model the item transitions, denoted as \IT, and their effects 
on the next baskets. 
%
Specifically, we model the item transitions via an encoder-decoder based framework, denoted as \transModel, which 
takes the individual items in the historical interactions as input to predict the items in the next baskets.
%

\paragraph{\transModel Encoder}
\label{sec:method:itemsTrans:encoder}

We first represent the items aggregated over all the baskets of each user using a vector \his:
\begin{equation}
\label{eqn:h}
\his = [g_1, g_2, \cdots, g_j, \cdots, g_n] \in \mathbb{R}^{1 \times n}, 
\end{equation}
where $g_{j}$ is the total number of interactions with item $j$ of the user among all her/his baskets, 
 weighted by a time-decay parameter: 
%
\begin{equation}
\label{eqn:gdecay}
    g_{j} = \sum\nolimits_{t=1}^{T} \gamma^{T-t} \mathbbm{1} (\text{item } j \in \Set(t)),
\end{equation}
where $\gamma \in (0, 1]$ is the time-decay parameter to emphasize the items in the most recent baskets more than 
those in early baskets, 
and $\mathbbm{1}(x)$ is an indicator function ($\mathbbm{1}(x) = 1$ if $x$ is true, otherwise 0). 
Existing methods usually use RNNs to learn weights for different baskets.
However, recommendation datasets are always super sparse so that  
RNNs may not learn meaningful weights in such sparse datasets.
Instead, in \methodAll, we leverage the fact that the recent interacted 
items affect the next basket of items more significantly 
compared to the items interacted much earlier~\cite{caser,sasrec}, and use the time-decay factor $\gamma$ to
explicitly assign and incorporate the different weights. 
%
%

Given $\his$, we use a simple fully-connected layer as the encoder to encode
 the hidden representation of the next basket $\mathbf{h}\in \mathbb{R}^{1\times d}$ as follows: 
%
\begin{equation}
\label{eqn:latent}
    \mathbf{h} = \tanh (\his W),
\end{equation}
where {$W  \in \mathbb{R}^{n \times d}$}
is a learnable weight matrix and $\tanh()$ is the non-linear hyperbolic tangent 
activation function.
%
Thus, the fully-connected layer represents the transition from all previous items 
to the items in the next basket. 
Here, we do not explicitly normalize $\his$ because 
the learnable parameter $W$ will accommodate the normalization.
Different from RNN-based methods which learn the transition patterns in a recurrent fashion 
and update the hidden states sequentially at each time stamp,  
the aggregation through the fully-connected layer in Equation~{\ref{eqn:latent}} can be done much more efficiently
as the item representation in Equation~\ref{eqn:h} can be done within a map-reduce framework~\cite{dean2004mapreduce} 
and thus in parallel. 
%
%
%
%
Therefore, \transModel could be more efficient than RNN-based methods especially on modeling 
long interaction sequences. 

\paragraph{\transModel Decoder}
\label{sec:method:itemsTrans:decoder}

%
Given $\mathbf{h}$, we use a fully-connected layer as the decoder to decode the recommendation scores $\mathbf{s}$ for all the item candidates in the next basket 
as follows:
\begin{equation}
\label{eqn:tcomb}
    \mathbf{s} = \text{softmax}(\mathbf{h}A + \mathbf{b}), 
\end{equation}
where $\mathbf{s}\in \mathbb{R}^{1\times n}$ is a vector in which the $j$-th dimension has 
the recommendation score of item $j$,
$A \in \mathbb{R}^{d \times n} $ is a learnable matrix 
and $\mathbf{b}$ is a bias vector.
%
The bias vector can also be interpreted as the 
items' global popularities because it is shared among all the baskets. 
Thus,  \transModel could capture both the transition patterns and items' global popularities.

\subsection{Calculating Recommendation Scores in \methodAll }
\label{sec:method:scores}

\subsubsection{Recommendation Scores using \GP and \IGP}
\label{sec:method:popularity}

We propose a variant of \methodAll to generate recommendations by combining the representations of 
users' general preferences $\mathbf{p}$ and items' global popularities $\mathbf{v}$ only. 
This method is referred to as mixed models with preferences and popularities
and denoted as \FREQ. 
In \FREQ, the recommendation scores of item candidates are calculated as follows:
\begin{equation}
\label{eqn:scoreFREQ}
\hat{\mathbf{r}} = (1-\alpha) \mathbf{p} +  \alpha \text{ softmax}(\mathbf{v}),
\end{equation}
where $\hat{\mathbf{r}} \in \mathbb{R}^{1 \times n}$ is the vector of recommendation scores, 
and $\alpha$ is a learnable weight to model the importance of  
users' general preferences and items' global popularities in users' interactions.
The softmax function is employed to normalize $\mathbf{v}$ to be 
in the same range with $\mathbf{p}$.
The intuition here is that, as shown in the literature~\cite{sets2sets,fpmc}, 
users' general preferences and items' global popularities significantly affect users' interactions.
Thus, combing these two important factors should lead to reasonable recommendations.
Based on the scores, the items with the top-$k$ largest  
scores will be recommended into the next basket.

In {\FREQ}, in principle, $\alpha$ could be modeled as a tunable parameter or a learnable weight.
To be consistent with the other {\methodAll} variants that will be presented in Section~\mbox{\ref{sec:method:gate}}
and Section~\mbox{\ref{sec:method:score}}, and to optimize performance, 
we model $\alpha$ as a learnable weight, and learn it in an end-to-end fashion.

\subsubsection{\mbox{Recommendation Scores using Gating Networks}}
\label{sec:method:gate}

One possible limitation of \FREQ could be that in \FREQ, we use a single weight $\alpha$ for all the users.
In this way, \FREQ can not capture the pattern that the weight could be different on different users.
To resolve this limitation,
we follow the idea of gating networks~\cite{hgn} 
to calculate personalized weight $\alpha$.
Specifically, we calculate the $\alpha$ using $\mathbf{p}$ (Equation~\ref{eqn:p}) and $\mathbf{v}$ (Equation~\ref{eqn:v}) 
as follows:
\begin{equation}
\label{eqn:weight}
    \alpha = \sigma(\mathbf{p}\RGateBeta + \mathbf{v}\PGateBeta),
\end{equation}
where $\sigma()$ is the sigmoid function, {\RGateBeta} and {\PGateBeta}  are learnable weight vectors.
The intuition here is that the importance of \GP and \IGP 
{(i.e., $\alpha$)} 
would be learned 
from themselves (i.e., $\mathbf{p}$ and $\mathbf{v}$).
The method with personalized weights is 
referred to as mixed models with gated preferences and popularities, 
denoted as \FREQP.

\subsubsection{\mbox{Recommendation Scores using \GP, \IGP and \IT}}
\label{sec:method:score}

Considering all the three important factors, we propose a unified method 
with preferences, popularities and transitions, denoted as \method. 
In \method we calculate the recommendation scores vector $\hat{\mathbf{r}}\in \mathbb{R}^{1\times n}$
using the representation $\mathbf{p}$ (Equation~\ref{eqn:p}) generated from \GP and the
recommendation scores $\mathbf{s}$ (Equation~\ref{eqn:tcomb})
from \transModel as follows:
\begin{equation}
\label{eqn:rec}
        \hat{\mathbf{r}} = (1 - \alpha) \mathbf{p} + \alpha \mathbf{s}, 
\end{equation}
where, similarly with that in \FREQP, 
$\alpha$ is calculated 
from $\mathbf{p}$ (Equation~\ref{eqn:p}) and $\mathbf{h}$ (Equation~\ref{eqn:latent}) as following: 
\begin{equation}
\label{eqn:weight}
    \alpha = \sigma(\mathbf{p}\RGateBeta + \mathbf{h}\PGateBeta),
    \end{equation}
where, as presented in Section~\ref{sec:method:gate}, 
$\sigma()$ is the sigmoid function, {\RGateBeta} and {\PGateBeta}  are learnable weight vectors.
Please note that as discussed in Section~\ref{sec:method:itemsTrans:decoder}, 
the scores in $\mathbf{s}$ are generated using both items' popularities and the transition patterns.
Thus, \method uses all the three factors to make recommendations.
Also note that, as shown in Equation~{\ref{eqn:tcomb}}, the vector $\mathbf{s}$ is already normalized to be in the same range with $\mathbf{p}$.
Therefore, we do not need the softmax function for the normalization in Equation~{\ref{eqn:rec}}.
%

\subsection{Network Training}
\label{sec:method:obj}

We minimize the negative log likelihood that the 
ground-truth items in the next baskets have high recommendation scores.
The optimization problem is formulated as follows, 
\begin{equation}
   \label{eqn:obj}
   \min\nolimits_{\boldsymbol{\Theta}} \sum\nolimits^{m}_{i=1} 
   -\mathbf{r}_i\log(\hat{\mathbf{r}}^{\mathsf{T}}_i)  + \lambda\|\boldsymbol{\Theta}\|^2,
\end{equation}
%
where $m$ is the number of users to recommend baskets to, 
$\mathbf{r}_i$ and $\hat{\mathbf{r}}_i$ are for the $i$-th user, 
$\boldsymbol{\Theta}$ is the set of the parameters, and $\lambda$ is the regularization parameter. 
Following previous work~\cite{dream,sets2sets}, we calculate the training error on the last basket in training data.
The vector $\mathbf{r}_i$ is the vector representation of the items in the last basket $\Set_i(T)$, 
in which the dimension $j$ is 1 if item $j$ is in $\Set_i(T)$ or 0 otherwise.
Here, we do not consider the frequencies of individual items in the baskets (i.e, $\mathbf{r}_i$ is binary),
as we do not predict the frequencies of items
in the next baskets. 
%
%
We optimize Problem~\ref{eqn:obj} using the Adagrad optimization method~\cite{adagrad}.
{The parameter tuning protocol and all the parameters for 
modeling are reported in the Supplementary Materials.



\section{Experimental Settings}
\label{sec:exp}

\subsection{Baseline Methods}
\label{sec:exp:baselines}

We compare \methodAll with 5 state-of-the-art baseline methods on next-basket recommendations: 
    1) \emph{\POP}~\cite{sets2sets} ranks items based on their popularity among all the users, 
    and recommends the top-$k$ most popular items. 
    %
    2) \emph{\POEP}~\cite{sets2sets} ranks items based on their popularity on each user and 
    recommends the personalized top-$k$ most popular items.
    3) \emph{\Dream}~\cite{dream} uses RNNs to 
    model users' preferences over time.
    It uses the most recent hidden state of RNNs to generate recommendation scores
    and recommends the items with top-$k$ scores.
    4) \emph{\FPMC}~\cite{fpmc} models users' long-term preferences and the transition patterns of items using 
    the first-order markov chain and matrix factorization.
    %
    5) \emph{\SetsSets}~\cite{sets2sets} adapts the encoder-decoder RNNs
    to model the short-term preferences and the recurrent behaviors of users. 
    %
 %
%
Please note that \SetsSets achieves the state-of-the-art performance on the next-basket 
recommendation and outperforms other methods~\cite{dream,fpmc,guo2003knn}.
Therefore, we compare \methodAll with \SetsSets 
but not  the methods that \SetsSets outperforms.

\subsection{Datasets}
\label{sec:exp:datasets}

We generate 4 datasets from 3 
benchmark datasets 
TaFeng\footnote{\mbox{\url{https://www.kaggle.com/chiranjivdas09/ta-feng-grocery-dataset}}}, 
TMall\footnote{\mbox{\url{https://tianchi.aliyun.com/dataset/dataDetail?dataId=42}\label{tafeng}}}, 
and Gowalla\footnote{\mbox{\url{https://snap.stanford.edu/data/loc-Gowalla.html}}} to evaluate the different methods.
TaFeng has grocery transactions in 4 months (i.e., 11/1/2020 to 02/28/2020) 
at a grocery store and each basket is a transaction of grocery items.
TMall has online transactions in 5 months (i.e., 07/01/2015 to 11/31/2015) 
and each basket is a transaction of products.
Gowalla~{\cite{cho2011friendship}} is a place-of-interests dataset
and contains user-venue check-in records with timestamps.
Similarly to Ying \etal~{\cite{shan}}, we view the check-in records in one day as a basket 
and focus on the records 
in 10 months (i.e., 01/01/2010 to 10/31/2010).

Following previous work~\cite{shan}, we do the following filtering to generate the datasets we will use in the experiments: 
1) filter out the infrequent users with
fewer than 10, 20 and 15 items from the original TaFeng, TMall and Gowalla dataset, respectively, 
2) filter out infrequent items interacted by
fewer than 10, 20 and 25 users from the TaFeng, TMall and Gowalla dataset, respectively, and
3) filter out users with fewer than 2 baskets.
Out of the above three filtering steps, each of the 3 original datasets will have frequent users and items, and we  
denote the processed datasets still as TaFeng, TMall and Gowalla. 
In order to better evaluate the methods in real applications that have a large amount of users and items, 
from the original TMall dataset, we also apply a smaller threshold 10 on user frequency and item frequency to 
generate another dataset, denoted as sTMall, with more users and items retained.
The statistics of the preprocessed datasets are presented in Table~\ref{tbl:datasets}. 
We noticed that the Dunnhumby dataset\footnote{\url{https://www.dunnhumby.com/source-files/}} 
and the Instacart dataset\footnote{\mbox{\url{https://www.instacart.com/datasets/grocery-shopping-2017}}} 
are also used in the literature~\mbox{\cite{adaloyal,sets2sets}}. 
However, Dunnhumby is a simulated dataset and the Instacart dataset is not publicly available now.
Therefore we do not use these datasets in our experiments.
We discussed the limitations of these datasets in detail in the Supplementary Materials.
\begin{table}[!t]
  \caption{Dataset Statistics}
  \label{tbl:datasets}
  \centering
  \begin{threeparttable}
      \begin{tabular}{
	@{\hspace{4pt}}l@{\hspace{4pt}}
	@{\hspace{4pt}}r@{\hspace{4pt}}
        @{\hspace{4pt}}r@{\hspace{4pt}}
        @{\hspace{4pt}}r@{\hspace{4pt}}
        @{\hspace{4pt}}r@{\hspace{4pt}}
        @{\hspace{4pt}}r@{\hspace{4pt}}
	}
        \toprule
        dataset & \#items & \#baskets & \#users & \#items/bskt & \#bskt/user\\
        \midrule
        TaFeng & 10,829 & 97,509 & 16,788 & 6.72 & 5.81\\
        TMall & 21,812 & 360,587 & 28,827 & 2.41 & 12.51\\
        sTMall & 104,266 & 2,052,959 & 214,105 & 2.01 & 9.59\\
        Gowalla & 26,529 & 902,505 & 26,822 & 1.77 & 33.65\\
        \bottomrule
      \end{tabular}
      \begin{tablenotes}
        \setlength\labelsep{0pt}
        \begin{footnotesize}
        \item
          The columns \#items, \#baskets, \#users, \#items/bskt and \#bskt/user correspond to
          the number of items, the number of baskets over all users, the number of users,
          the average number of items per basket and the average number of baskets per user, respectively.
          \par
        \end{footnotesize}
      \end{tablenotes}
  \end{threeparttable}
  \vspace{-10pt}
\end{table}


\subsection{Experimental Protocol}
\label{sec:exp:proto}

Similarly to Ying \etal~\cite{shan}, we split the 4 datasets 
based on cut-off times as shown in Figure~\ref{fig:split}. 
Specifically, on TaFeng, 
we use the transactions in the first 3 months as the training set, 
the transactions in the following 0.5 month as the validation set, 
and the transactions in the last 0.5 month as the testing set.
Similarly, on Gowalla,
we use the records in the first 8 months as the training set, 
the records in the following 1 month as the validation set, 
and the records in the last 1 month as the testing set.
On TMall and sTMall, 
we use the transactions in the first 3.5 months as the training set, 
the transactions in the following 0.5 month as the validation set, and the transactions in the last 1 month as the testing set.
We split the datasets in this way to guarantee that all the interactions in the testing set occur
after the interactions in the training and validation sets. Thus, the setting is close to real use scenarios.  
A detailed discussion about different experimental protocols is presented later in Section~{\ref{sec:dis:setting}.}

%

We denote the baskets in the training, validation and testing sets as training, validation 
and testing baskets, respectively.
The users which have interactions in the training, validation and testing sets are denoted
as training, validation and testing users, respectively.
Please note that a user can be both training and testing user if she/he has baskets 
in both training and testing sets.
During training, we only use the interactions in the training baskets 
to estimate users' general preferences and to learn item transition patterns.
There could be items in testing or validation baskets that never appeared in training baskets (i.e. cold-start items).
In this case, we will retain the baskets with such items. 
Since \methodAll and all the baseline methods are not developed for the cold-start problem~\cite{coldStrat}, 
the cold-start items will not get recommended but the baskets with such items can still be evaluated due to other items. 

We tune the parameters using grid search and use the best parameters 
in terms of recall@$5$ on the validation set during testing 
for the \methodAll and all the baseline methods.
Following previous work~\cite{caser,sasrec,hgn}, during testing, 
we use the interactions in both training and validation sets to train the model with the optimal parameters identified 
at the validation set.
Similarly to Hu \etal~\cite{sets2sets}, we evaluate \methodAll and baseline methods on three tasks:
recommending the first next basket, the second next basket and the third next basket.
Please note that in recommending the second next or third next basket, during evaluation, 
the first or second testing basket, respectively, of testing users will be used to update 
the user's general preference representation $\mathbf{p}$ (Equation~\ref{eqn:p}) and 
item transitions in $\mathbf{g}$ (Equation~\ref{eqn:h}).
%
%
%
Also note that the number of validation and testing users in these three tasks could be different. 
In recommending the second next basket, only users with at least two validation 
or testing baskets are used as validation or testing users, but 
users with only one validation or testing basket will not be used in evaluation.


\begin{figure}[!t]
    \centering
    \includegraphics[width=\linewidth,]{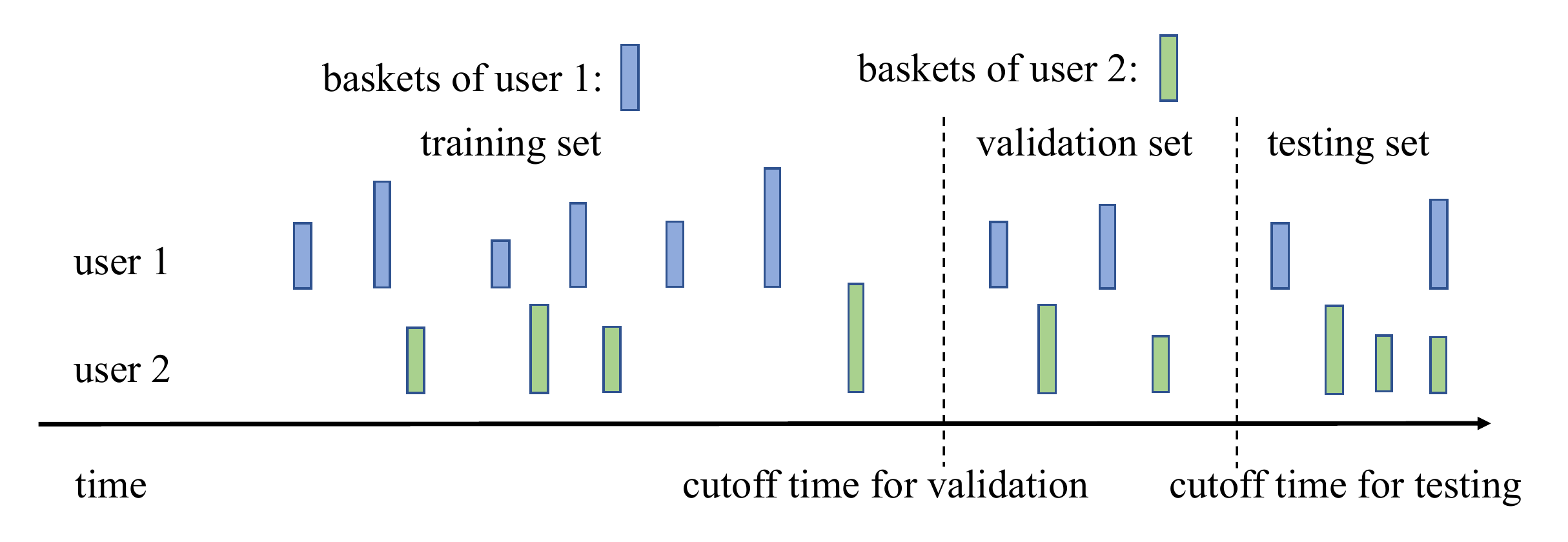}
    \caption{\method Datasets Splitting Protocol}
    \label{fig:split}
    \vspace{-15pt}
\end{figure}

\subsection{Evaluation Metrics}
\label{sec:exp:metric}

We use recall@$k$, precision@$k$, and NDCG@$k$ to evaluate the different methods. 
For each user, recall measures the proportion of all the ground-truth interacted items 
in a testing basket that are correctly recommended.  
We denote the set of $k$ recommended items and the set of the items in the ground-truth basket
as $R_k$ and $S$, respectively.
Given $R_k$ and $S$,  recall@$k$ is calculated as follows:
\begin{equation}
    \label{eqn:recall}
    \text{recall@}k = \frac{|R_k \cap S|}{|S|},
\end{equation}
where $R_k \cap S$ is the intersection between the two sets and $|S|$ denotes 
the size of the set $S$.
Precision measures the proportion of all the recommended items that are correctly recommended, 
and precision@$k$ is calculated as follows:

\begin{equation}
    \label{eqn:precision}
\text{precision@}k = \frac{|R_k \cap S|}{k}.
\end{equation}
We report in the experimental results the recall@$k$ and precision@$k$ values that are calculated as the average over all the testing users. 
Higher recall@$k$ and precision@$k$ indicate better performance. 
It is worth noting that although we use precision@$k$ in our experiments, we argue that precision@$k$ 
may not be a proper metric for 
evaluating next-basket recommendation methods
as we will discuss later in Section~{\ref{sec:dis:metrics}}.

NDCG@$k$ 
is the normalized discounted cumulative gain for the top-$k$ ranking.  
In our experiments, the gain indicates whether a ground-truth item is recommended 
(i.e., gain is 1) or not (i.e., gain is 0). 
NDCG@$k$ incorporates the positions of the correctly recommended
items among the top-k recommendations. 
Higher NDCG@$k$ indicates the ground-truth items are recommended at very top, and thus better recommendation performance. 
%

Besides these evaluation metrics, 
we also statistically test the significance of the performance difference among different methods via a standard $t$-test. 
Specifically, 
we conducted \mbox{$t$-test} over the paired recall, 
NDCG and precision values from different methods. 
If the $p$-values are smaller than a predefined threshold $\alpha$ ($\alpha$ = 0.05 in our experiments), 
the performance difference of two methods is considered statistically significant at 100(1-$\alpha$)\% confidence level.


\section{Experimental Results}
\label{sec:results}

\subsection{Overall Performance on the First Next Basket}
\label{sec:results:performanceNextOne}

\begin{table}[!h]
  \caption{Performance Comparison on the Next Basket}
  \label{tbl:performanceNextOne}
  \centering
  \begin{threeparttable}
      \begin{tabular}{
	@{\hspace{0pt}}l@{\hspace{5pt}}
	@{\hspace{4pt}}l@{\hspace{0pt}}          
	@{\hspace{4pt}}r@{\hspace{4pt}}
        @{\hspace{4pt}}r@{\hspace{4pt}}
        @{\hspace{4pt}}r@{\hspace{4pt}}
        @{\hspace{4pt}}r@{\hspace{4pt}}
        @{\hspace{4pt}}r@{\hspace{4pt}}
        @{\hspace{4pt}}r@{\hspace{4pt}}
        @{\hspace{4pt}}r@{\hspace{4pt}}
        @{\hspace{4pt}}r@{\hspace{0pt}}
	}
        \toprule
        \multirow{2}{*}{} & \multirow{2}{*}{method} & \multicolumn{3}{c}{recall@$k$} 
        && \multicolumn{3}{c}{NDCG@$k$}\\
        \cmidrule(lr){3-5} \cmidrule(lr){7-9} 
        & & $k$=5 & $k$=10 & $k$=20 && $k$=5 & $k$=10 & $k$=20\\
        \midrule
        \multirow{9}{*}{\rotatebox[origin=c]{90}{TaFeng (7,227)}}
        & \POP        & \underline{0.0866} & 0.0963 & 0.1151 && \underline{0.1227} & \underline{0.1161} & 0.1203\\
        & \POEP       & 0.0817 & 0.1153 & 0.1563 && 0.1109 & 0.1127 & \underline{0.1240}\\
        & \Dream      & 0.0839 & 0.0928 & 0.1086 && 0.0694 & 0.0655 & 0.0697\\
        & \FPMC       & 0.0568 & 0.0672 & 0.0831 && 0.0691 & 0.0658 & 0.0698\\
        & \SetsSets   & 0.0822 & \underline{0.1230} & \underline{0.1705} && 0.0952 & 0.1049 & 0.1200\\
        & \FREQ        & 0.0908 & 0.1338 & 0.1766 && 0.1192 & 0.1244 & 0.1367\\
        & \FREQP       & 0.0916 & 0.1344 & 0.1782 && 0.1207 & 0.1257 & 0.1381\\
        & \method     & \textbf{$\mathclap{^{\dagger~}}$0.1013} & \textbf{$\mathclap{^{\dagger~}}$0.1375} & \textbf{$\mathclap{^{\dagger~}}$0.1936} 
        && \textbf{$\mathclap{^{\dagger~}}$0.1280} & \textbf{$\mathclap{^{\dagger~}}$0.1306} & \textbf{$\mathclap{^{\dagger~}}$0.1469}\\
        \cmidrule(lr){2-9}
        & improv     & 17.0\%$\mathclap{^*}$ & 11.8\%$\mathclap{^*}$  & 13.5\%$\mathclap{^*}$ && 4.3\% & 12.5\%$\mathclap{^*}$ & 18.5\%$\mathclap{^*}$\\
        \midrule
        \multirow{9}{*}{\rotatebox[origin=c]{90}{TMall (14,051)}}
        & \POP       & 0.0802 & 0.0828 & 0.0872 && 0.0777 & 0.0784 & 0.0800\\
        & \POEP      & 0.1051 & 0.1264 & 0.1524 && 0.0793 & 0.0857 & 0.0927\\
        & \Dream     & 0.0833 & 0.0868 & 0.0927 && 0.0752 & 0.0765 & 0.0781\\
        & \FPMC      & 0.0802 & 0.0809 & 0.0867 && 0.0777 & 0.0778 & 0.0797\\
        & \SetsSets  & \underline{0.1092} & \underline{0.1360} & \underline{$\mathclap{^{\dagger~}}$0.1653} 
        && \underline{$\mathclap{^{\dagger~}}$0.0979} & \underline{$\mathclap{^{\dagger~}}$0.1071} & \underline{$\mathclap{^{\dagger~}}$0.1154}\\
        & \FREQ       & 0.1118 & 0.1365 & 0.1584 && 0.0843 & 0.0919 & 0.0977\\
        & \FREQP      & 0.1123 & 0.1360 & 0.1548 && 0.0846 & 0.0919 & 0.0971\\
        & \method    & \textbf{$\mathclap{^{\dagger~}}$0.1165} & \textbf{$\mathclap{^{\dagger~}}$0.1395} & \textbf{0.1648} 
        && \textbf{0.0939} & \textbf{0.1010} & \textbf{0.1079}\\
        \cmidrule(lr){2-9}
        & improv     & 6.7\%$\mathclap{^*}$  & 2.6\%$\mathclap{^*}$  & -0.3\% && -4.1\%$\mathclap{^*}$  & -5.7\%$\mathclap{^*}$  & -6.5\%$\mathclap{^*}$ \\
        \midrule
        \multirow{9}{*}{\rotatebox[origin=c]{90}{sTMall (94,337)}}
        & \POP       & 0.0859 & 0.0880 & 0.0905 && \underline{0.0834} & \underline{0.0840} & 0.0846\\
        & \POEP      & \underline{0.0936} & \underline{0.1091} & \underline{0.1187} && 0.0761 & 0.0810 & 0.0836\\
        & \Dream     & 0.0852 & 0.0873 & 0.0934 && 0.0826 & 0.0833 & \underline{0.0848}\\
        & \FPMC      & 0.0845 & 0.0869 & 0.0902 && 0.0820 & 0.0828 & 0.0837\\
        & \SetsSets  & OOM    & OOM    & OOM  && OOM    & OOM    & OOM\\
        & \FREQ       & 0.0991 & 0.1203 & 0.1388 && 0.0791 & 0.0857 & 0.0906\\
        & \FREQP     & 0.0992 & 0.1204 & 0.1393 && 0.0791 & 0.0857 & 0.0907\\
        & \method    & \textbf{$\mathclap{^{\dagger~}}$0.1114} & \textbf{$\mathclap{^{\dagger~}}$0.1285} & \textbf{$\mathclap{^{\dagger~}}$0.1404} 
        && \textbf{$\mathclap{^{\dagger~}}$0.0948} & \textbf{$\mathclap{^{\dagger~}}$0.1002} & \textbf{$\mathclap{^{\dagger~}}$0.1035}\\
        \cmidrule(lr){2-9}
        & improv     & 19.0\%$\mathclap{^*}$  & 17.8\%$\mathclap{^*}$  & 18.3\%$\mathclap{^*}$  && 13.7\%$\mathclap{^*}$  & 19.3\%$\mathclap{^*}$  & 22.1\%$\mathclap{^*}$ \\
        \midrule
        \multirow{9}{*}{\rotatebox[origin=c]{90}{Gowalla (12,975)}}
        & \POP       & 0.0111 & 0.0240 & 0.0413 && 0.0064 & 0.0110 & 0.0158\\
        & \POEP      & \underline{0.4551} & \underline{0.5179} & \underline{0.5649} && \underline{0.3793} & \underline{0.4007} & \underline{0.4136}\\
        & \Dream     & 0.0187 & 0.0307 & 0.0436 && 0.0127 & 0.0169 & 0.0206\\
        & \FPMC      & 0.0107 & 0.0255 & 0.0536 && 0.0059 & 0.0111 & 0.0187\\
        & \SetsSets  & 0.3941 & 0.4745 & 0.5443 && 0.3184 & 0.3462 & 0.3654\\
        & \FREQ       & 0.4574 & 0.5213 & 0.5664 && 0.3800 & 0.4019 & 0.4143\\
        & \FREQP     & 0.4578 & 0.5194 & 0.5689 && 0.3802 & 0.4013 & 0.4148\\
        & \method    & \textbf{$\mathclap{^{\dagger~}}$0.4599} & \textbf{$\mathclap{^{\dagger~}}$0.5232} & \textbf{$\mathclap{^{\dagger~}}$0.5736} 
        && \textbf{$\mathclap{^{\dagger~}}$0.3813} & \textbf{$\mathclap{^{\dagger~}}$0.4030} & \textbf{$\mathclap{^{\dagger~}}$0.4168}\\
        \cmidrule(lr){2-9}
        & improv     & 1.1\%$\mathclap{^*}$  & 1.0\%$\mathclap{^*}$  & 1.5\%$\mathclap{^*}$  && 0.5\%$\mathclap{^*}$  & 0.6\%$\mathclap{^*}$  & 0.8\%$\mathclap{^*}$ \\
        \bottomrule
      \end{tabular}
      \begin{tablenotes}
        \setlength\labelsep{0pt}
	\begin{footnotesize}
	\item
        {For each dataset, the best performance among our proposed methods (i.e., {\FREQ}, {\FREQP} and {\method}) is in {\textbf{bold}}, 
        the best performance among the baseline methods is {\underline{underlined}}, 
        and the overall best performance is indicated by a dagger (i.e., {${\dagger~}$}).
        The row "improv" presents the percentage improvement
	of  the best performing methods among {\footnotesize{\FREQ}}, {\footnotesize{\FREQP}} and {\footnotesize{\method}} ({\textbf{bold}}) over the best performing baseline 
	methods ({\underline{underlined}}) in each column.}
	The numbers in the parentheses after the datasets represent the number of testing users in the datasets.
	The "OOM" represents the out of memory issue.
	{The ${^*}$ indicates that the improvement is statistically significant at 95 percent confidence level.}
          \par
	\end{footnotesize}
      \end{tablenotes}
  \end{threeparttable}
  \vspace{-5pt}
\end{table}




Table~\ref{tbl:performanceNextOne} presents the overall performance at recall@$k$ and NDCG@$k$
in recommending the first next basket
of all the methods on the 4 datasets. 
Due to the space limit, we report the performance at precision@$k$ in the Supplementary Materials.
In Table~{\ref{tbl:performanceNextOne}}, for each dataset, the best performance among {\methodAll} variants (i.e., {\FREQ}, {\FREQP} and {\method}) is in bold, 
the best performance among baseline methods (e.g., {\POP}, {\POEP}, {\SetsSets}) is underlined.
and the overall best performance is indicated by a dagger (i.e., $^\dagger$).
We report the parameters that achieve the reported performance also in the Supplementary Materials.
%
For \SetsSets, we use the implementation provided by the authors. 
However, this implementation raises memory issues 
and cannot fit in 16GB GPU memory
on the largest dataset sTMall. 
Therefore, we report out of memory (OOM) for \SetsSets 
on sTMall. 

Table~\ref{tbl:performanceNextOne} shows that 
overall, \method is the best performing method on the task of recommending the first next basket.
%
%
In terms of recall@$5$, recall@$10$ and recall@$20$, 
\method achieves the best performance with significant 
improvement compared to the second best method on TaFeng and sTMall.
On the TMall and Gowalla datasets, \method also achieves the best or second best performance 
at recall@$5$, recall@$10$ and recall@$20$.
Compared to the second best method, {\method} achieves on average 6.8\%, 2.9\% and 2.5\% improvement at recall@$5$, recall@$10$ and recall@$20$, respectively, 
over all the datasets.
In terms of NDCG@$5$, NDCG@$10$ and NDCG@$20$, 
\method achieves the best performance on TaFeng, sTMall and Gowalla, and 
the second best performance on the TMall dataset.
In particular, on the largest dataset sTMall, \method achieves substantial improvement of 10.8\% on average over 
all the metrics compared to the second best method. 
On the most widely used benchmark dataset TaFeng, \method also achieves significant improvement of at least 2.3\% 
over the second best method at all the metrics. 
On Gowalla where many baseline methods do not perform well, \method is still slightly better than the second best method 
\FREQP. 
It is also worth noting that, compared to the performance of the best baseline methods (underlined in Table~{\ref{tbl:performanceNextOne}}), 
{\method} achieves statistically significant improvement over most of the metrics on 3 out of 4 datasets.
On the TMall dataset, {\method} still achieves statistically significant improvement over the best baseline methods at both recall@$5$ and recall@$10$.
These results demonstrate the strong performance of \method.
\FREQP is the second best performing method in our experiments.
It achieves the second best or (near) the second best performance on all the four datasets. 
We notice that \POP, \Dream and \FPMC work poorly on the Gowalla dataset.
This might be due to the fact that these methods do not really capture the personalized general 
preferences of users.
Recall that the Gowalla dataset is a place-of-interests dataset, which 
contains user-venue check-in records of users.
Different users live in different places and could interact with very different items. 
Thus, methods that do not explicitly model users' personalized general preferences
could not work well on this dataset.
%

\subsubsection{Comparing \method with model-based methods}
\label{sec:results:performanceNextOne:methodWmodel}

Table~\ref{tbl:performanceNextOne} also shows that among the 4 model-based methods 
\Dream, \FPMC, \SetsSets and \method, 
\method consistently and significantly outperforms \Dream and \FPMC on all the datasets.
The primary difference among \method, \Dream and \FPMC is 
that \method explicitly models users' general preferences using the frequencies of the items 
that each user has interactions with, while \Dream and \FPMC implicitly model 
them using the hidden state of RNNs or user embeddings.
Given the sparse nature of recommendation datasets (Table~\ref{tbl:datasets}), it is possible that 
the learned hidden states or user embeddings cannot represent the user preferences well, as the 
signals of user preferences are smoothed out due to data sparsity 
during the recurrent updates, or by the pooling or weighting schemes
used to learn user embeddings as some other work also noticed~\cite{ham,jain2019attention,dacrema2019we}. 
The superior performance of \method over \Dream and \FPMC on all the datasets demonstrates 
the effect of explicitly modeling users' general preferences.

Table~\ref{tbl:performanceNextOne} shows that 
\method significantly outperforms \SetsSets on all the datasets except TMall in terms of 
both recall@$k$ and NDCG@$k$. 
The primary differences between \method and \SetsSets are 
1) \method explicitly models the transition patterns among items using encoder-decoder-based \transModel,
while \SetsSets implicitly models the transition patterns using RNNs, and
2) when calculating the recommendation scores, 
\method learns a single weight on each user (i.e., $\alpha$ in Equation~\ref{eqn:rec}),
but \SetsSets learns different weights for different items on each user. 
%
Given the sparse nature of the recommendation datasets, weights 
for different items on each user may not be well learned~\cite{ham,jain2019attention}.
Thus, such weights may not necessarily 
help better differentiate user general preferences over items.
In addition, the learned weights over items may guide the model to learn 
inaccurate general preferences of users, and thus degrade the performance. 
%
We also notice that on TMall, \method underperforms \SetsSets in terms of NDCG but 
outperforms \SetsSets in terms of recall.
This indicates that on certain datasets, \method could be more effective than \SetsSets on ranking the 
items of users' interest  on top of the recommendation list, while less effective than 
\SetsSets on raking these items on the very top.
However, \SetsSets is very memory consuming, demonstrated by out of memory (OOM) issues on the largest dataset sTMall, which 
substantially limits its use in real, large-scale recommendation problems.

\subsubsection{Comparing \method with popularity-based methods}
\label{sec:results:performanceNextOne:methodWpop}

In Table~\ref{tbl:performanceNextOne}, 
we also notice that \method statistically significantly outperforms the best popularity-based method \FREQP 
on all the datasets.
On average, it achieves  6.8\%, 3.0\%, 9.3\% and 7.8\% improvement over \FREQP in terms of 
recall@5, recall@10, NDCG@5 and NDCG@10, respectively, over all the datasets.
Recall that the key difference between \method and \FREQP is that \method models users' general preferences, 
items' global popularities and the transition patterns, 
whereas \FREQP only models users' general preferences 
and items' global popularities. 
These results demonstrate the importance of transition patterns 
in sequence-based next-basket recommendation.

\subsubsection{Comparison among popularity-based methods}
\label{sec:results:performanceNextOne:AmongPop}

Among the four popularity-based methods \POP, \POEP, \FREQ and \FREQP, 
\FREQP achieves the best performance at most of the metrics on all the 4 datasets.
Between \FREQP and \FREQ, 
\FREQP outperforms \FREQ on the TaFeng and Gowalla datasets, and achieves 
similar performance with \FREQ on the TMall and sTMall datasets. 
In terms of recall@$5$, \FREQP outperforms \FREQ on all the datasets.
In terms of recall@$10$ and recall@$20$, \FREQP outperforms \FREQ on the TaFeng and sTMall datasets, 
and achieves similar performance with \FREQ on the TMall and Gowalla datasets.
We also found a similar trend on NDCG@$k$:
for example, in terms of NDCG@$5$,
\FREQP outperforms \FREQ on all the datasets except sTMall.
On sTMall, \FREQP achieves the same performance with \FREQ. 
The difference between \FREQP and \FREQ is that \FREQP learns personalized weights to 
combine users' general preferences and items' global popularities, 
while \FREQ only learns one such weight for all the users. 
The substantial performance improvement of \FREQP over \FREQ 
demonstrates the importance of learning personalized weights.
We also notice that overall, \FREQ consistently outperforms \POP and \POEP on all the datasets over all the metrics.
In terms of recall@$5$, recall@$10$ and recall@$20$, \FREQ consistently outperforms 
\POP and \POEP at all the 4 datasets. 
For example, on the widely used TaFeng dataset, in terms of recall@$10$, \FREQ achieves significant improvement of 38.9\% and 16.6\% compared to \POP and \POEP, respectively.
%
Recall that the difference between \FREQ, \POP and \POEP is that \FREQ models both 
users' general preferences and items' global popularities, 
while \POP and \POEP only model one of them.
The substantial improvement of \FREQ over \POP and \POEP demonstrates that items' global popularities and users' general preferences are complementary. 
When learned together, they will enable better performance than each alone.
It is also worth noting that \FREQP outperforms the state-of-the-art model-based method \SetsSets 
at all the metrics on TaFeng and Gowalla.
The superior performance of \FREQP is a strong evidence that the 
simple popularity-based methods could still be very effective 
in next-basket recommendations. 

%
\subsection{Performance on the Second Next Basket}
\label{sec:results:performanceNextTwo}

\begin{table}
  \caption{Performance on the Second Next Basket}
  \label{tbl:performanceNextTwo}
  \centering
  \begin{threeparttable}
      \begin{tabular}{
	@{\hspace{0pt}}l@{\hspace{5pt}}
	@{\hspace{4pt}}l@{\hspace{0pt}}          
	@{\hspace{4pt}}r@{\hspace{4pt}}
        @{\hspace{4pt}}r@{\hspace{4pt}}
        @{\hspace{4pt}}r@{\hspace{4pt}}
        @{\hspace{4pt}}r@{\hspace{4pt}}
        @{\hspace{4pt}}r@{\hspace{4pt}}
        @{\hspace{4pt}}r@{\hspace{4pt}}
        @{\hspace{4pt}}r@{\hspace{4pt}}
        @{\hspace{4pt}}r@{\hspace{0pt}}
	}
        \toprule
         \multirow{2}{*}{} & \multirow{2}{*}{method} & \multicolumn{3}{c}{recall@$k$} 
        && \multicolumn{3}{c}{NDCG@$k$}\\
        \cmidrule(lr){3-5} \cmidrule(lr){7-9} 
        & & $k$=5 & $k$=10 & $k$=20 && $k$=5 & $k$=10 & $k$=20\\
        \midrule
        \multirow{9}{*}{\rotatebox[origin=c]{90}{TaFeng (2,801)}}
        & \POP       & \underline{0.1024} & \underline{0.1352} & 0.1475 
        && \underline{$\mathclap{^{\dagger~}}$0.1356} & \underline{$\mathclap{^{\dagger~}}$0.1392} & \underline{0.1422}\\ 
        & \POEP     & 0.0920 & 0.1313 & 0.1787 && 0.1056 & 0.1138 & 0.1293\\ 
        & \Dream   & 0.0965 & 0.1054 & 0.1168 && 0.0629 & 0.0619 & 0.0651\\
        & \FPMC     & 0.0461 & 0.0618 & 0.0805 && 0.0476 & 0.0500 & 0.0558\\
        & \SetsSets & 0.0734 & 0.1236 & \underline{0.1882} && 0.0670 & 0.0856 & 0.1086\\
        & \FREQ      & 0.0893 & 0.1367 & 0.1927 && 0.1053 & 0.1161 & 0.1345\\
        & \FREQP     & \textbf{$\mathclap{^{\dagger~}}$0.1113} & \textbf{$\mathclap{^{\dagger~}}$0.1549} & 0.2036 
        && \textbf{0.1222} & \textbf{0.1316} & \textbf{$\mathclap{^{\dagger~}}$0.1482}\\
        & \method   & \textbf{$\mathclap{^{\dagger~}}$0.1113} & 0.1517 & \textbf{$\mathclap{^{\dagger~}}$0.2062} && 0.1198 & 0.1280 & 0.1461\\
        \cmidrule(lr){2-9}
        & improv    & 8.7\% & 14.6\%$\mathclap{^*}$ & 9.6\%$\mathclap{^*}$ && -9.9\%$\mathclap{^*}$ & -5.5\% & 4.2\%\\
        \midrule
        \multirow{9}{*}{\rotatebox[origin=c]{90}{TMall (5,109)}}
        & \POP      & 0.0855 & 0.0872 & 0.0892 && 0.0827 & 0.0837 & 0.0844\\
        & \POEP     & 0.1253 & 0.1556 & 0.1904 && 0.0959 & 0.1052 & 0.1144\\
        & \Dream    & 0.0907 & 0.0940 & 0.0979 && 0.0844 & 0.0857 & 0.0868\\
        & \FPMC     & 0.0860 & 0.0875 & 0.0915 && 0.0831 & 0.0837 & 0.0849\\
        & \SetsSets & \underline{0.1345} & \underline{0.1628} & \underline{0.1972} 
        && \underline{$\mathclap{^{\dagger~}}$0.1175} & \underline{$\mathclap{^{\dagger~}}$0.1275} & \underline{$\mathclap{^{\dagger~}}$0.1373}\\
        & \FREQ      & 0.1344 & \textbf{$\mathclap{^{\dagger~}}$0.1657} & 0.1940 && 0.1019 & 0.1117 & 0.1192\\
        & \FREQP     & 0.1347 & 0.1645 & \textbf{$\mathclap{^{\dagger~}}$0.2018} && 0.1022 & 0.1112 & 0.1211\\
        & \method   & \textbf{$\mathclap{^{\dagger~}}$0.1374} & 0.1656 & 0.1999 
        && \textbf{0.1096} & \textbf{0.1183} & \textbf{0.1276}\\
        \cmidrule(lr){2-9}
        & improv    & 2.2\% & 1.8\% & 2.3\%$\mathclap{^*}$ && -6.7\%$\mathclap{^*}$ & -7.2\%$\mathclap{^*}$ & -7.1\%$\mathclap{^*}$\\
        \midrule
        \multirow{9}{*}{\rotatebox[origin=c]{90}{sTMall (29,741)}}
        & \POP      & 0.0835 & 0.0870 & 0.0912 && 0.0820 & 0.0832 & 0.0843\\ 
        & \POEP     & \underline{0.1132} & \underline{0.1398} & \underline{0.1563} && \underline{0.0885} & \underline{0.0968} & \underline{0.1011}\\ 
        & \Dream    & 0.0866 & 0.0893 & 0.0946 && 0.0833 & 0.0842 & 0.0856\\
        & \FPMC     & 0.0853 & 0.0874 & 0.0911 && 0.0828 & 0.0835 & 0.0844\\
        & \SetsSets & OOM    & OOM    & OOM    && OOM    & OOM    & OOM\\
        & \FREQ      & 0.1205 & 0.1482 & 0.1698 && 0.0926 & 0.1012 & 0.1069\\
        & \FREQP     & 0.1203 & 0.1482 & 0.1699 && 0.0925 & 0.1012 & 0.1069\\
        & \method   & \textbf{$\mathclap{^{\dagger~}}$0.1258} & \textbf{$\mathclap{^{\dagger~}}$0.1528} & \textbf{$\mathclap{^{\dagger~}}$0.1718}  
        && \textbf{$\mathclap{^{\dagger~}}$0.1023} & \textbf{$\mathclap{^{\dagger~}}$0.1109} & \textbf{$\mathclap{^{\dagger~}}$0.1159}\\
        \cmidrule(lr){2-9}
        & improv    & 11.1\%$\mathclap{^*}$ & 9.3\%$\mathclap{^*}$ & 9.9\%$\mathclap{^*}$ && 15.6\%$\mathclap{^*}$ & 14.6\%$\mathclap{^*}$ & 14.6\%$\mathclap{^*}$\\
        \midrule
        \multirow{9}{*}{\rotatebox[origin=c]{90}{Gowalla (10,032)}}
        & \POP       & 0.0124 & 0.0228 & 0.0399 && 0.0072 & 0.0110 & 0.0158\\
        & \POEP     & \underline{0.4765} & \underline{0.5413} & \underline{0.5872} && \underline{0.3920} & \underline{0.4142} & \underline{0.4271}\\
        & \Dream    & 0.0200 & 0.0340 & 0.0507 && 0.0134 & 0.0182 & 0.0228\\
        & \FPMC     & 0.0059 & 0.0158 & 0.0329 && 0.0033 & 0.0067 & 0.0112\\
        & \SetsSets & 0.3915 & 0.4804 & 0.5565 && 0.3128 & 0.3436 & 0.3646\\
        & \FREQ      & 0.4764 & 0.5426 & 0.5894 && 0.3916 & 0.4145 & 0.4275\\
        & \FREQP    & 0.4767 & 0.5439 & 0.5904 && 0.3921 & 0.4152 & 0.4281\\
        & \method   & \textbf{$\mathclap{^{\dagger~}}$0.4787} & \textbf{$\mathclap{^{\dagger~}}$0.5456} & \textbf{$\mathclap{^{\dagger~}}$0.5979} 
        && \textbf{$\mathclap{^{\dagger~}}$0.3932} & \textbf{$\mathclap{^{\dagger~}}$0.4163} & \textbf{$\mathclap{^{\dagger~}}$0.4307}\\ 
        \cmidrule(lr){2-9}
        & improv    & 0.5\% & 0.8\%$\mathclap{^*}$ & 1.8\%$\mathclap{^*}$ && 0.3\% & 0.5\%$\mathclap{^*}$ & 0.8\%$\mathclap{^*}$\\
        \bottomrule
      \end{tabular}
      \begin{tablenotes}
        \setlength\labelsep{0pt}
	\begin{footnotesize}
	\item
	The columns in this table have the same meanings as those in Table~\ref{tbl:performanceNextOne}.
          \par
	\end{footnotesize}
      \end{tablenotes}
  \end{threeparttable}
  \vspace{-10pt}
\end{table}



Table~\ref{tbl:performanceNextTwo} presents the overall performance of different methods at recall@$k$ and NDCG@$k$ 
in recommending the second next basket (i.e., the second basket in the testing set) on the 4 datasets. 
We also report the performance at precision@$k$ in the Supplementary Materials.
The parameter tuning protocol in this task is the same as that in recommending 
the first next basket (Section~\ref{sec:results:performanceNextOne}).
As discussed in Section~\ref{sec:exp:proto}, when recommending the second next basket, 
the first testing basket of users will be used to update the models. 
In addition, in this task, only users with at least two testing baskets will be used as testing users.
Thus, the number of testing users in this task could be different from that in recommending the first next basket.
Specifically, as shown in Table~\ref{tbl:performanceNextTwo},  
when recommending the second next basket, we have 2,801, 5,109, 29,741 
and 10,032 testing users on TaFeng, TMall, sTMall and Gowalla, respectively.

\subsubsection{Overall Performance}
\label{exp:compWfirst}

As shown in Table~\ref{tbl:performanceNextTwo}, overall, in recommending the second next basket, 
the performance of \methodAll and baseline methods has a similar trend 
as that in recommending the first next basket.
In particular, \method is still the best performing method in this task. 
In terms of recall@$5$, \method achieves the best performance on all the 4 datasets.
In terms of recall@$10$, \method achieves the best performance on the sTMall and Gowalla datasets, 
and the second best performance on the TaFeng and TMall datasets.
We also found a similar trend on NDCG@$k$:
in terms of NDCG@$5$, 
\method achieves the best performance on the sTMall and Gowalla datasets, 
and the second best or (near) the second best performance on the TaFeng and TMall datasets.
\FREQP is still the second best performing method.
In terms of recall@$5$ and recall@$10$, \FREQP achieves the best performance on the TaFeng dataset, 
and the second best performance or (near) the second best performance on the other 3 datasets 
(i.e., TMall, sTMall, Gowalla).
In terms of  NDCG@$5$ and NDCG@$10$, \FREQP also achieves the second best 
or (near) the second best performance  on 3 out of 4 datasets (i.e., TaFeng, sTMall, Gowalla).
It is also worth noting that on the widely used TaFeng dataset, 
\FREQP significantly outperforms \FREQ at 24.6\%, 13.3\%, 16.0\% 
and 13.4\% on recall@$5$, recall@$10$, 
NDCG@$5$ and NDCG@$10$, respectively.
As discussed in Section~\ref{sec:results:performanceNextOne:AmongPop}, the difference between 
\FREQP and \FREQ is that \FREQP learns personalized combine weights, 
while \FREQ learns one combine weight for all the users.
The significant improvement of \FREQP over \FREQ further demonstrates the importance 
of learning personalized combine weights.

\subsubsection{Comparing with the performance on the next basket}
\label{sec:results:performanceNextTwo:compWfirst}

%
We also notice that the performance of those methods that model users' general preferences 
(e.g., \POEP, \FREQP and \method)
increases as we 
recommend the baskets in the later future (i.e., the second next basket).
For example, on the largest sTMall dataset, \POEP has recall@$5$ value 0.0936 (Table~\ref{tbl:performanceNextOne}) 
in recommending the first next basket, 
while this value increases to 0.1132 (Table~\ref{tbl:performanceNextTwo}) 
in recommending the second next basket.
This might be due to the fact that the testing users with more than one basket in the testing set 
are in general more active (i.e., have more baskets).
Specifically, on sTMall, the testing users in the experiments of recommending the first next basket 
have 9.6 baskets on average used for training.
However, the testing users in the experiments of recommending the second next basket 
have 10.6 
baskets on average (i.e. 10.4\% increasing).
Thus, more baskets used for model training 
enable methods which model users' general preferences 
to more accurately estimate 
the general preferences of testing users, and thus achieve better performance for the second next basket recommendation.
%

It is worth noting that although \FREQP significantly underperforms 
\method when recommending the first next basket, 
\FREQP could achieve similar 
or even better performance over \method at some metrics when recommending the second next basket.
For example, on TaFeng, when recommending the first next basket, 
\method achieves significant improvement of 10.6\%, 8.6\% at recall@$5$ and recall@$20$ (Table~\ref{tbl:performanceNextOne}) over \FREQP.
However, when recommending the second next basket, \FREQP is able to achieve 
the same performance with \method 
at recall@$5$ (i.e., 0.1113 as in Table~\ref{tbl:performanceNextTwo}).
As just discussed, the testing users in 
recommending the second next basket are in general more active than those in 
recommending the first next basket.
The similar performance of \FREQP and \method indicates that the interactions of active users 
are more dominated by their general preferences and the global popularities of items.  
Thus, for active users, the simple popularity-based methods could be very effective.
However, since in real applications, most of the users are not active, it is still important to model 
the transition patterns in general recommendation applications.

\subsection{Performance on the Third Next Basket}
\label{sec:results:performanceNextThree}

\begin{table}
  \caption{Performance on the Third Next Basket}
  \label{tbl:performanceNextThree}
  \centering
  \begin{threeparttable}
      \begin{tabular}{
	@{\hspace{0pt}}l@{\hspace{4pt}}
	@{\hspace{4pt}}l@{\hspace{0pt}}          
	@{\hspace{4pt}}r@{\hspace{4pt}}
        @{\hspace{4pt}}r@{\hspace{4pt}}
        @{\hspace{4pt}}r@{\hspace{4pt}}
        @{\hspace{4pt}}r@{\hspace{4pt}}
        @{\hspace{4pt}}r@{\hspace{4pt}}
        @{\hspace{4pt}}r@{\hspace{4pt}}
        @{\hspace{4pt}}r@{\hspace{4pt}}
        @{\hspace{4pt}}r@{\hspace{0pt}}
	}
        \toprule
        \multirow{2}{*}{} & \multirow{2}{*}{method} & \multicolumn{3}{c}{recall@$k$} 
        && \multicolumn{3}{c}{NDCG@$k$}\\
        \cmidrule(lr){3-5} \cmidrule(lr){7-9} 
        & & $k$=5 & $k$=10 & $k$=20 && $k$=5 & $k$=10 & $k$=20\\
        \midrule
        \multirow{9}{*}{\rotatebox[origin=c]{90}{TaFeng (1,099)}}
        & \POP       & 0.0725 & 0.1255 & 0.1519 && 0.0924 & 0.1063 & 0.1142\\
        & \POEP      & \underline{0.1037} & \underline{0.1415} & \underline{0.1907} && \underline{0.1114} & \underline{0.1207} & \underline{0.1360}\\
        & \Dream     & 0.0632 & 0.0763 & 0.0866 && 0.0552 & 0.0569 & 0.0601\\
        & \FPMC      & 0.0420 & 0.0593 & 0.0788 && 0.0451 & 0.0499 & 0.0562\\
        & \SetsSets  & 0.0732 & 0.1158 & 0.1791 && 0.0650 & 0.0802 & 0.1025\\
        & \FREQ       & 0.1039 & 0.1482 & 0.1906 && 0.1109 & 0.1231 & 0.1367\\
        & \FREQP      & \textbf{$\mathclap{^{\dagger~}}$0.1162} & \textbf{$\mathclap{^{\dagger~}}$0.1547} & \textbf{$\mathclap{^{\dagger~}}$0.2057} 
        && \textbf{$\mathclap{^{\dagger~}}$0.1205} & \textbf{$\mathclap{^{\dagger~}}$0.1297} & \textbf{$\mathclap{^{\dagger~}}$0.1461}\\
        & \method    & 0.1141 & 0.1525 & 0.1969 && 0.1162 & 0.1273 & 0.1421\\
        \cmidrule(lr){2-9}
        & improv     & 12.1\%$\mathclap{^*}$ & 9.3\%$\mathclap{^*}$ & 7.9\%$\mathclap{^*}$ && 8.2\%$\mathclap{^*}$ & 7.5\%$\mathclap{^*}$ & 7.4\%$\mathclap{^*}$\\
        \midrule
        \multirow{9}{*}{\rotatebox[origin=c]{90}{TMall (1,461)}}
        & \POP       & 0.0727 & 0.0741 & 0.0759 && 0.0675 & 0.0678 & 0.0682\\
        & \POEP      & \underline{0.1522} & \underline{0.1925} & 0.2308 && 0.1096 & 0.1209 & 0.1306\\
        & \Dream     & 0.0717 & 0.0744 & 0.0799 && 0.0655 & 0.0663 & 0.0677\\
        & \FPMC      & 0.0736 & 0.0756 & 0.0791 && 0.0682 & 0.0684 & 0.0696\\
        & \SetsSets  & 0.1512 & 0.1898 & \underline{0.2368} 
        && \underline{$\mathclap{^{\dagger~}}$0.1256} & \underline{$\mathclap{^{\dagger~}}$0.1387} & \underline{$\mathclap{^{\dagger~}}$0.1517}\\
        & \FREQ       & 0.1568 & 0.1942 & 0.2348 && 0.1135 & 0.1247 & 0.1348\\
        & \FREQP     & 0.1586 & \textbf{$\mathclap{^{\dagger~}}$0.1965} & 0.2320 && 0.1150 & \textbf{0.1260} & 0.1349\\
        & \method     & \textbf{$\mathclap{^{\dagger~}}$0.1603} & 0.1909 & \textbf{$\mathclap{^{\dagger~}}$0.2390} && \textbf{0.1152} & 0.1238 & \textbf{0.1358}\\
        \cmidrule(lr){2-9}
        & improv     & 5.3\%$\mathclap{^*}$ & 2.1\% & 0.9\% && -8.3\%$\mathclap{^*}$ & -9.2\%$\mathclap{^*}$ & -10.5\%$\mathclap{^*}$\\
        \midrule
        \multirow{9}{*}{\rotatebox[origin=c]{90}{sTMall (7,561)}}
        & \POP       & 0.0802 & 0.0824 & 0.0854 && 0.0781 & 0.0788 & 0.0795\\
        & \POEP      & \underline{0.1267} & \underline{0.1610} & \underline{0.1872} 
        && \underline{0.0984} & \underline{0.1087} & \underline{0.1155}\\
        & \Dream     & 0.0838 & 0.0864 & 0.0903 && 0.0800 & 0.0808 & 0.0819\\
        & \FPMC      & 0.0824 & 0.0846 & 0.0884 && 0.0792 & 0.0800 & 0.0808\\
        & \SetsSets  & OOM    & OOM    & OOM    && OOM    & OOM    & OOM\\
        & \FREQ       & 0.1348 & \textbf{$\mathclap{^{\dagger~}}$0.1705} & \textbf{$\mathclap{^{\dagger~}}$0.1961} 
        && 0.1029 & 0.1139 & 0.1205\\
        & \FREQP     & 0.1352 & 0.1703 & 0.1960 && 0.1030 & 0.1137 & 0.1204\\
        & \method     & \textbf{$\mathclap{^{\dagger~}}$0.1373} & 0.1696 & 0.1954 
        && \textbf{$\mathclap{^{\dagger~}}$0.1088} & \textbf{$\mathclap{^{\dagger~}}$0.1187} & \textbf{$\mathclap{^{\dagger~}}$0.1254}\\
        \cmidrule(lr){2-9}
        & improv     & 8.4\%$\mathclap{^*}$ & 5.9\%$\mathclap{^*}$ & 4.8\%$\mathclap{^*}$ && 10.6\%$\mathclap{^*}$ & 9.2\%$\mathclap{^*}$ & 8.6\%$\mathclap{^*}$\\
        \midrule
        \multirow{9}{*}{\rotatebox[origin=c]{90}{Gowalla (7,985)}}
        & \POP       & 0.0108 & 0.0233 & 0.0402 && 0.0062 & 0.0107 & 0.0155\\
        & \POEP      & \underline{0.5092} & \underline{0.5751} & \underline{0.6251} && \underline{0.4282} & \underline{0.4509} & \underline{0.4649}\\
        & \Dream     & 0.0187 & 0.0295 & 0.0442 && 0.0127 & 0.0166 & 0.0208\\
        & \FPMC      & 0.0179 & 0.0404 & 0.0789 && 0.0094 & 0.0172 & 0.0274\\
        & \SetsSets  & 0.4367 & 0.5230 & 0.5981 && 0.3496 & 0.3799 & 0.4009\\
        & \FREQ       & 0.5137 & 0.5779 & 0.6282 && 0.4299 & 0.4518 & 0.4657\\
        & \FREQP      & 0.5133 & \textbf{$\mathclap{^{\dagger~}}$0.5802} & 0.6268 && 0.4296 & 0.4525 & 0.4654\\
        & \method    & \textbf{$\mathclap{^{\dagger~}}$0.5154} & \textbf{$\mathclap{^{\dagger~}}$0.5802} & \textbf{$\mathclap{^{\dagger~}}$0.6321} 
        && \textbf{$\mathclap{^{\dagger~}}$0.4309} & \textbf{$\mathclap{^{\dagger~}}$0.4531} & \textbf{$\mathclap{^{\dagger~}}$0.4675}\\
        \cmidrule(lr){2-9}
        & improv     & 1.2\%$\mathclap{^*}$ & 0.9\%$\mathclap{^*}$ & 1.1\%$\mathclap{^*}$ && 0.6\%$\mathclap{^*}$ & 0.5\%$\mathclap{^*}$ & 0.6\%$\mathclap{^*}$\\
        \bottomrule 
      \end{tabular}
      \begin{tablenotes}
        \setlength\labelsep{0pt}
	\begin{footnotesize}
	\item
	The columns in this table have the same meanings as those in Table~\ref{tbl:performanceNextOne}.
          \par
	\end{footnotesize}
      \end{tablenotes}
  \end{threeparttable}
  \vspace{-10pt}
\end{table}


Table~\ref{tbl:performanceNextThree} presents the overall performance of methods at recall@$k$ and NDCG@$k$
on the task of recommending the third next basket.
The performance at precision@$k$ is reported in the Supplementary Materials.
%
Please note that as discussed in Section~\ref{sec:exp:proto} 
and Section~\ref{sec:results:performanceNextTwo}, 
the number of testing users in this task could be different from that in recommending 
the first, and second next basket.
Specifically, as shown in Table~\ref{tbl:performanceNextThree},  
when recommending the third next basket, we have 1,099, 1,461, 7,561 
and 7,985 testing users on TaFeng, TMall, sTMall and Gowalla, respectively.
Table~\ref{tbl:performanceNextThree} shows that overall, the performance of 
\methodAll and baseline methods still has similar trend 
as that in recommending the first and second next basket.
\method is still the best performing method. 
In terms of recall@$5$, \method achieves the best performance at 3 out of 4 datasets 
(i.e., TaFeng, sTMall and Gowalla).
On the TMall dataset, \method also achieves the second best performance.
We also found a similar trend on NDCG@$k$:
in terms of NDCG@$5$, \method also achieves 
the best performance on 3 out of 4 datasets, and the second best performance on the TMall dataset.
\FREQP is still the second best performing method.
In terms of recall@$5$, \FREQP achieves the best performance on the TMall dataset, and the second best performance on the TaFeng and sTMall dataset.
The same trends as discussed in Section~\ref{exp:compWfirst} could also be found here. 
%
It is also worth noting that as shown in Table~{\ref{tbl:performanceNextThree}}, in terms of recall@$k$, 
the best {\methodAll} variant (i.e., {\FREQ}, {\FREQP} or {\method}), 
statistically significantly outperforms the best baseline methods on 3 out of 4 datasets. 
On TMall, {\method} still statistically significantly outperforms the best baseline method {\POEP} at recall@$5$.

\subsection{Performance Summary among All the Tasks}
\label{exp:summary}
%

Table~\ref{tbl:performanceNextOne}, Table~\ref{tbl:performanceNextTwo} and Table~\ref{tbl:performanceNextThree} together show that 
\method is the best performing method over all the 3 tasks. 
It significantly outperforms the state-of-the-art baseline method \SetsSets at all the metrics 
over all the 3 tasks.
For example, in terms of recall@$5$, \method achieves 15.5\%, 25.4\% and 26.6\% improvement 
on average over all the datasets except sTMall in recommending the first, second and third next basket, respectively.
These results demonstrate the strong ability of \method in next-basket recommendation.
Table~\ref{tbl:performanceNextOne}, Table~\ref{tbl:performanceNextTwo} and Table~\ref{tbl:performanceNextThree} together also show that 
\FREQP achieves the second best performance over the 3 tasks.
It is worth noting that although \FREQP does not perform as well as \method, 
it still consistently outperforms the state-of-the-art baseline method \SetsSets over all the 3 tasks.
These results demonstrate the strong effectiveness of simple 
popularity-based methods in next-basket recommendation.

\subsection{Ablation Study}
\label{sec:results:ablation}

\subsubsection{Comparing Different Factors in \method}
\label{exp:ablation:factor}

\begin{table}
  \caption{Ablation Study on the Next Basket}
  \label{tbl:ablationOne}
  \centering
  \begin{threeparttable}
      \begin{tabular}{
	@{\hspace{1pt}}l@{\hspace{4pt}}
	@{\hspace{5pt}}l@{\hspace{4pt}}          
	@{\hspace{4pt}}r@{\hspace{4pt}}
        @{\hspace{4pt}}r@{\hspace{4pt}}
        @{\hspace{4pt}}r@{\hspace{4pt}}
        @{\hspace{4pt}}r@{\hspace{4pt}}
        @{\hspace{4pt}}r@{\hspace{4pt}}
        @{\hspace{4pt}}r@{\hspace{4pt}}
        @{\hspace{4pt}}r@{\hspace{1pt}}
	}
        \toprule
        \multirow{2}{*}{} & \multirow{2}{*}{method} & \multicolumn{3}{c}{recall@$k$}
        && \multicolumn{3}{c}{NDCG@$k$}\\
        \cmidrule(lr){3-5} \cmidrule(lr){7-9}
        & & $k$=5 & $k$=10 & $k$=20 && $k$=5 & $k$=10 & $k$=20\\
        \midrule
        \multirow{3}{*}{\rotatebox[origin=c]{90}{TaFeng}}
        & \GP & \underline{0.0817} & \underline{0.1153} & \underline{0.1563} 
        && \underline{0.1109} & \underline{0.1127} & \underline{0.1240}\\ 
        & \IT & 0.0508 & 0.0774 & 0.1129 && 0.0660 & 0.0701 & 0.0807\\
        & \method & \textbf{0.1013} & \textbf{0.1375} & \textbf{0.1936} 
        && \textbf{0.1280} & \textbf{0.1306} & \textbf{0.1469}\\
        \midrule
        \multirow{3}{*}{\rotatebox[origin=c]{90}{TMall}}
        & \GP & \underline{0.1051} & \underline{0.1264} & \underline{0.1524} 
        && 0.0793 & 0.0857 & \underline{0.0927}\\
        & \IT & 0.0947 & 0.1045 & 0.1162 
        && \underline{0.0851} & \underline{0.0880} & 0.0915\\
        & \method & \textbf{0.1165} & \textbf{0.1395} & \textbf{0.1648} 
        && \textbf{0.0939} & \textbf{0.1010} & \textbf{0.1079}\\
        \midrule
        \multirow{3}{*}{\rotatebox[origin=c]{90}{sTMall}}
        & \GP & \underline{0.0936} & \underline{0.1091} & \underline{0.1187} 
        && 0.0761 & 0.0810 & 0.0836\\
        & \IT & 0.0928 & 0.0983 & 0.1052 && \underline{0.0856} & \underline{0.0873} & \underline{0.0892}\\
        & \method & \textbf{0.1114} & \textbf{0.1285} & \textbf{0.1404} 
        && \textbf{0.0948} & \textbf{0.1002} & \textbf{0.1035}\\
        \midrule
        \multirow{3}{*}{\rotatebox[origin=c]{90}{Gowalla}}
        & \GP & \underline{0.4551} & \underline{0.5179} & \underline{0.5649} 
        && \underline{0.3793} & \underline{0.4007} & \underline{0.4136}\\
        & \IT & 0.3105 & 0.3342 & 0.3567 && 0.2778 & 0.2856 & 0.2917\\
        & \method & \textbf{0.4599} & \textbf{0.5232} & \textbf{0.5736} 
        && \textbf{0.3813} & \textbf{0.4030} & \textbf{0.4168}\\
        \bottomrule
      \end{tabular}
      \begin{tablenotes}
        \setlength\labelsep{0pt}
        \begin{footnotesize}
        \item
        \method is identical to $\GP$+$\IT$. 
          The best and second best performance in each dataset is in \textbf{bold} and \underline{underlined}, respectivley. 
          \par
        \end{footnotesize}
      \end{tablenotes}
  \end{threeparttable}
\end{table}


We conduct an ablation study to verify the 
effects of the different components (i.e., \GP, \IT) in \method. 
%
We present the next basket recommendation results generated 
by \GP and \IT alone, and their combination \method
in Table~\ref{tbl:ablationOne}. 
Note that \GP recommends the personalized most popular items to each user, and thus it is identical to \POEP. 
When testing \GP, the final recommendation scores $\hat{\mathbf{r}}$ (Equation~\ref{eqn:rec}) 
are identical to those based on users' general preferences in $\mathbf{p}$ (Equation~\ref{eqn:p}) (i.e., 
$\alpha$=$0$ in Equation~\ref{eqn:rec}).
When testing \IT, essentially it is to test \transModel and the final recommendation scores are 
in $\mathbf{s}$ (Equation~\ref{eqn:tcomb}) (i.e., $\alpha$=$1$ in Equation~\ref{eqn:rec}).

Table~\ref{tbl:ablationOne} shows that \GP is a strong baseline for all the methods on all the datasets. This indicates 
the importance of users' general preferences in the next-basket recommendation. 
%
%
\IT does not outperform \GP in terms of recall@$k$ on all the datasets.
We also found a similar trend on NDCG@$k$.  
In terms of NDCG@$k$, \GP significantly outperforms \IT on TaFeng and Gowalla 
and achieves similar performance with \IT on the TMall and sTMall datasets.
%
When \IT is combined with \GP (i.e., \method in Table~\ref{tbl:ablationOne}),
there is a notable increase compared to each individual \IT and \GP. 
This may be because that in \method, as \GP captures the general preferences, 
\IT can learn the remaining, transition patterns and items' global popularities that cannot be captured by \GP.
%
%
In Table~\ref{tbl:ablationOne}, 
\method (i.e., $\GP$+$\IT$) achieves the best performance on all the 4 datasets.
It also shows improvement from $\GP$ and $\IT$.
This indicates that when learned together, \GP and \IT are complementary and 
enable better performance than each alone.

\subsubsection{Comparing \transModel and RNN-based Methods}
\label{sec:results:ablation:rnn}

We also notice that as shown in Table~\ref{tbl:performanceNextOne} and Table~\ref{tbl:ablationOne}, 
\transModel (i.e., the model to learn \IT), an encoder-decoder based approach (Section~\ref{sec:method:itemsTrans}), 
on its own outperforms \Dream (i.e., RNN-based method) on 3 out of 4 datasets 
i.e., TMall, sTMall and Gowalla) 
at both recall@$k$ and NDCG@$k$,
and achieves comparable results with \Dream on the TaFeng dataset at NDCG@$k$.
For example, on TMall, \transModel achieves 0.0947 in terms of recall@$5$ (Table~\ref{tbl:ablationOne}) compared to \Dream 
with 0.0833 (Table~\ref{tbl:performanceNextOne}), that is, \transModel is  13.7\% better than \Dream.
Similarly, in terms of 
recall@$10$ and recall@$20$, \transModel achieves 0.1045 and 0.1162 (Table~\ref{tbl:ablationOne}), respectively, 
compared to \Dream with 0.0868 and 0.0927 (Table~\ref{tbl:performanceNextOne}), respectively, 
that is, \transModel achieves 20.4\% improvement at recall@$10$ and 25.4\% improvement at recall@$20$ compared to \Dream. 
We also found a similar trend on sTMall.
In terms of recall@$5$,  \transModel achieves 8.9\% improvement over \Dream (0.0928 vs 0.0852) on sTMall.
These results are strong evidence to show that \transModel 
\noindent
could outperform
RNN-based methods on benchmark datasets.
It is worth noting that as shown in Table~\ref{tbl:performanceNextOne} and Table~\ref{tbl:ablationOne}, 
on Gowalla, \transModel achieves reasonable performance, while \Dream fails.
As discussed in Section~\ref{sec:results:performanceNextOne}, for doing good recommendations on Gowalla, 
models should be able to learn users' general preferences from the interactions.
The reasonable and poor performance of \transModel and \Dream, respectively, 
indicates that \transModel could implicitly learn users' general preferences, 
while RNN-based methods might not.
We also notice that \transModel on its own does not work as well as \SetsSets as 
shown in Table~\ref{tbl:performanceNextOne} and Table~\ref{tbl:ablationOne}.
However, this might be due to the reason that \SetsSets models both the transition patterns 
and users' general preferences, 
while \transModel does not explicitly model users' general preferences.
When \transModel learned with \GP together (i.e, \method), \method outperforms 
\SetsSets on all the datasets as shown in Table~\ref{tbl:performanceNextOne}.
%
These results indicate that \transModel could be more effective than the RNNs used in \SetsSets 
on modeling transition patterns.

%
%
%

\subsection{Analysis on Transition Patterns}
\label{sec:results:weights}
%
\begin{figure}[!t]
  \hspace{-60pt}
  \begin{minipage}{\linewidth}
  \begin{center}
      \input{legend2.tex}
      \end{center}
  \end{minipage}
\\
        \begin{minipage}{0.24\textwidth}
               \centering
                \input{TaFeng_gate.tex}
                \vspace{13pt}
                \caption*{(a) TaFeng}
                \label{fig:TaFeng_gate}
        \end{minipage}
        \begin{minipage}{0.24\textwidth}
    
                \centering
                \input{TMall_gate.tex}
                \vspace{13pt}
                \caption*{(b) TMall}
                \label{fig:TMall_gate}
        \end{minipage}
\\
        \begin{minipage}{0.24\textwidth}
                \centering
                \input{TMalls_gate.tex}
                \vspace{13pt}
                \caption*{(c) sTMall}
                \label{fig:TMalls_gate}
        \end{minipage}
        \begin{minipage}{0.24\textwidth}
                \centering
                \input{Gowalla_gate.tex}
                \vspace{13pt}
                \caption*{(d) Gowalla}
                \label{fig:Gowalla_gate}
        \end{minipage}
\vspace{-5pt}
\caption{Distributions of Gating Weights from \method}
\label{fig:gate_plot}
\vspace{-10pt}
\end{figure}

%
%
%
%
%

We further analyze if \method learns good weights $\alpha$ (Equation~\ref{eqn:rec}) 
to differentiate the importance of \GP and \IT. 
Figure~\ref{fig:gate_plot} presents the distribution of the 
weights $\alpha$ from the best performing 
\method models on the 4 datasets.
Please note that as presented in Section~\ref{sec:method:factors:general},  
only the items interacted by the user will get non-zero 
recommendation scores in \GP,
while all the items could get non-zero recommendation scores in \IT.
As a result, for items with non-zero scores, the scale of their scores might be different in \GP and \IT. 
And thus, the absolute value of the weights on different components may not necessarily represent the 
true importance of the corresponding factors in users' behavior. 
For example, on TMall, users have higher weights on \IT than that on \GP. 
It does not necessarily indicate that the transition patterns are more important than users' general preference for the recommendation on this dataset.

As shown in Figure~\ref{fig:gate_plot}, on Gowalla, users' weights on \GP are much higher 
than that on the other datasets. 
This is consistent with the observation that  
on Gowalla, users' general preferences 
play a more important role for recommendation than that on the other datasets
(shown in Table~\ref{tbl:performanceNextOne}).
This consistency demonstrates that \method is able to learn good weights to differentiate the 
importance of \GP and \IT on different datasets and application scenarios.
%
%
%
%

\subsection{Cluster Analysis}
\label{sec:results:clusters}

\begin{figure}
   \centering
    \input{item_embedding.tex}
    \caption{Item embeddings from \method (TaFeng)}
    \label{fig:embedding}
    \vspace{-10pt}
\end{figure}

We further evaluate if \method
really learns the transition patterns among items. 
Specifically, we learn the weight matrix $W$ (Equation~{\ref{eqn:latent}}) in {\method} 
using the training and validation baskets in the widely used TaFeng dataset on 
recommending the first next basket, 
and export the matrix for the analysis.
Note that the weight matrix $W$ could be viewed as an item embedding matrix, 
in which each row is the embedding of a single item.
Given $W$, we evaluate if items with similar transition patterns will have similar embeddings.
To get the ground-truth transition patterns among items, 
we construct a matrix $T$ also from the training and validation baskets in TaFeng.
In $T$, $T_{ij}$ is the 
number of times that item $i$ in the previous baskets 
transits to item $j$ in the next basket. 
That is, the $i$-th row of $T$ contains the items 
that item $i$ has transited to.
Thus, $T$ contains the ground-truth transition patterns among items.
After constructing matrix $T$, 
we could get the items which have similar transition patterns by calculating the pairwise similarities.

%
Figure~\ref{fig:embedding}, generated using the t-SNE~\cite{tsne} method, presents the item embeddings 
generated from \method on the TaFeng dataset.
Specifically, we project the item embeddings in $W$ to the two-dimensional (2d) space using t-SNE, 
and then plot the projected embeddings of items in this figure.
%
In Figure~\ref{fig:embedding}, there are many well-formed clusters (e.g., $C_1$, $C_2$).
We find that generally, the items within the same cluster have similar transition patterns.
For example, the average pairwise similarity of items in $C_1$ and $C_2$ 
is 25.7\% and 11.4\% higher than that over all the item pairs, respectively.
These results demonstrate that the encoder-decoder framework (\transModel) in \method 
could effectively capture the transition patterns among items.

\subsection{Analysis of Diversity of Recommendations}
\label{sec:results:pop}

We also evaluate the diversity of the recommendations from different methods.
Due to the space limit, we report the results in the Supplementary Materials.
%
Generally, we find that {\method} could generate more diverse 
recommendations over all the baseline methods except {\POEP}.
Considering both the quality and diversity of the recommendations, 
{\method} significantly outperforms all the baseline methods, 
and could achieve superior performance in real applications.

\section{Discussions}
\label{sec:discusion}


\subsection{Experimental Protocols}
\label{sec:dis:setting}

A commonly used experimental protocol in the literature~\cite{sets2sets} is as follows. 
%
Users are randomly split into 5 or 10 folds to conduct 5 or 10-fold cross validation.
For each user in the testing fold, her/his last basket in sequential order is used as the testing basket, 
the other baskets are used as the training baskets.
For each user in the training folds, her/his last basket is used to measure training errors,
the other baskets are used to train the model and generate recommendation scores for the last basket.
%
When absolute time information is absent in the datasets, this experimental protocol enables full separation 
among the training and testing sets, and approximates real application scenario for each testing user. 
However, when the absolute time information is present, which is the case in most
of the popular benchmark datasets including TaFeng, TMall and Gowalla, 
this protocol will create artificial use scenario that deviates from that in real applications. 
The issue is that following this protocol, a basket in the training set from one user may have a later timestamp 
than a basket in the testing set from another user, and therefore a later basket is used to train a model to recommend 
an earlier basket, which is not realistic. 
Our protocol splits the training, validation and testing sets based on an absolute cut-off time for all the users, and thus 
avoids the above issue and is closer to real application scenarios. 
%
Another widely used experimental protocol~\mbox{\cite{dream,fpmc,ham,caser,dt4sr}} 
is that for each user,
her/his last and second last basket were used as the testing basket and validation basket, respectively;  
the other baskets are used as the training baskets.
This protocol has the same issue as discussed above.
Here, we refer this protocol as the order-based split protocol.
We evaluate {\methodAll} and baseline methods using this widely used but questionable order-based split protocol,
%
and report the results in the Supplementary Materials.
We found that, under the order-based split protocol, {\methodAll}
still achieves superior performance over the best baseline methods on all the datasets over most of the evaluation metrics.

Another commonly used experimental setting~\cite{dream,fpmc} is to evaluate different methods in 
recommending the first next basket. However, in real applications, the model is usually updated weekly or monthly, 
and thus would need to recommend multiple baskets for active users before model updates.
In this case, the performance in recommending the first next basket may not accurately represent the
models' effectiveness in real applications.
In our experiments, we also evaluate methods in the task of recommending a few next baskets to
more accurately and comprehensively evaluate the model performance in real applications.
%

\subsection{Evaluation Metrics}
\label{sec:dis:metrics}

In the experiments, we use recall@$k$ and NDCG@$k$ to evaluate different methods. 
These two metrics are important and widely used for top-$N$ recommendation~\cite{charu2016}, and also popular in 
sequential recommendations~\cite{sasrec,dt4sr,ham} and next-basket recommendations~\cite{fpmc,dream,sets2sets}. 
Recall@$k$ measures the proportion of all the ground-truth interacted items in a testing basket that are also  
among top-$k$ recommended items. We believe this is a proper metric to use because in the end, the 
recommendation methods aim to identify all the items that the users will be interested in eventually, that is, to maximize
recall. 
In addition, recall values at different top-$k$ positions also indicate the ranking structures of recommended items, 
where we prefer the items that users are interested in are ranked on top. 
NDCG also measures the ranking positions of the items that users are interested in.  Higher NDCG@$k$ values indicate
that more users' interested items are ranked on top.  
Since in real applications the users will look at a subset of the recommendations from the top 
of the recommendation list, we believe that evaluation metrics that consider ranking positions 
are more useful and applicable in real applications, and as discussed in Aggarwal~\cite{charu2016} (Chapter 7.5.5), 
NDCG is more suitable than ROC measures or rank-correlation coefficients in 
distinguishing between higher-ranked and lower-ranked items. 

%

%
%
%
The metric precision@$k$ is also a popular metric in evaluating recommendations. 
This metric, however, may not be proper for next-basket recommendation evaluation. 
First of all, precision@$k$ does not consider the ranking positions of the correctly recommended items. 
Second, the value of precision@$k$ is ``not necessarily monotonic in $k$ because both the numerator
and denominator may change with $k$ differently", as discussed in Aggarwal~\cite{charu2016} (Chapter 7.5.4). 
In addition, precision@$k$ could be strongly biased by basket sizes: for small baskets, 
precision@$k$ could be small even if all the 
items are correctly recommended. For example, 
if all the items in a size-2 basket are correctly recommended, precision@$10$ is only 0.2. 
However, for large baskets, precision@$k$ can be large even only a small portion of the
items are correctly recommended. 
For example, if 5 items of a size-20 basket are correctly recommended, that is, only 25\%
of the items are correctly recommended, precision@$10$ is 0.5. 
When only considering precision@$k$, we may prefer the second recommendation, even though it 
is half way to its best possible results (i.e., correctly recommend 10 among top-10 recommended items, 
with precision@10=1.0), 
but the first recommendation has already achieved its best possible results.  
Recall@$k$ alleviates such issues with a normalization using basket size. 
Therefore, precision and other precision-based metrics (e.g., AUC, F1)
may not be proper for evaluating 
next-basket recommendation methods.
However, to be comprehensive, we still use this metric in our experiments and report the results in the Supplementary Materials.
\section{Conclusions}
\label{sec:conclusion}

In this paper, we presented novel \methodAll models that 
conduct next-basket recommendation using three important 
factors: 1) users' general preferences, 2) items' global popularities 
and 3) the transition patterns among items. 
Our experimental results in comparison with 5 state-of-the-art 
next-basket recommendation methods on 4 public
benchmark datasets demonstrate substantial performance improvement from \methodAll 
in both the next basket recommendation (improvement of up to 19.0\% at recall@5) 
and the next a few baskets recommendation (improvement of up to 14.4\% at recall@5).
Our ablation study demonstrates
the importance of users' general preferences in next-basket recommendations, 
and the complementarity among all the factors in \methodAll. 
Our ablation study also demonstrates that the simple encoder-decoder based framework \transModel
(Section~\ref{sec:method:itemsTrans}) is more effective than RNNs 
for modeling the transition patterns
in benchmark datasets (improvement as much as 20.4\% at recall@5). 
Our analysis on the learned item embedding matrix further demonstrates 
that {\transModel}
could effectively capture the ground-truth transition patterns among items.

One potential limitation of {\methodAll} and the other data-driven basket recommendation methods 
is that the recommended items may not form realistic baskets.
For example, the method may recommend ten brands of milk as a basket to users.
However, in practice, users rarely purchase together ten brands in one basket.
To mitigate this potential limitation without sacrificing recommendation performance, 
we may need to carefully balance the modeling of item complementarities (additional discussions in the Supplementary Materials)
and the other important factors.
We leave the investigation of this problem in our future work.
In addition to this limitation, another future direction could be to extend {\methodAll} for the cold-start problem. 
We also leave the investigation of this problem as in our future work.

\section*{Acknowledgements}

This project was made possible, in part, by support from the National Science Foundation 
under Grant Number IIS-1855501, EAR-1520870, SES-1949037, IIS-1827472 and IIS-2133650, 
and from National Library of Medicine under Grant Number 1R01LM012605-01A1 and R21LM013678-01. 
Any opinions, findings, and conclusions or recommendations expressed 
in this material are those of the authors and do not necessarily reflect the views of the funding agencies.

\bibliographystyle{IEEEtran}
\bibliography{paper}

\begin{thebibliography}{10}
\providecommand{\url}[1]{#1}
\csname url@samestyle\endcsname
\providecommand{\newblock}{\relax}
\providecommand{\bibinfo}[2]{#2}
\providecommand{\BIBentrySTDinterwordspacing}{\spaceskip=0pt\relax}
\providecommand{\BIBentryALTinterwordstretchfactor}{4}
\providecommand{\BIBentryALTinterwordspacing}{\spaceskip=\fontdimen2\font plus
\BIBentryALTinterwordstretchfactor\fontdimen3\font minus
  \fontdimen4\font\relax}
\providecommand{\BIBforeignlanguage}[2]{{%
\expandafter\ifx\csname l@#1\endcsname\relax
\typeout{** WARNING: IEEEtran.bst: No hyphenation pattern has been}%
\typeout{** loaded for the language `#1'. Using the pattern for}%
\typeout{** the default language instead.}%
\else
\language=\csname l@#1\endcsname
\fi
#2}}
\providecommand{\BIBdecl}{\relax}
\BIBdecl

\bibitem{sets2sets}
H.~Hu and X.~He, ``Sets2sets: Learning from sequential sets with neural
  networks,'' in \emph{Proceedings of the 25th ACM SIGKDD International
  Conference on Knowledge Discovery and Data Mining}, 2019, pp. 1491--1499.

\bibitem{fpmc}
S.~Rendle, C.~Freudenthaler, and L.~Schmidt-Thieme, ``Factorizing personalized
  markov chains for next-basket recommendation,'' in \emph{Proceedings of the
  19th international conference on World wide web}, 2010, pp. 811--820.

\bibitem{dream}
F.~Yu, Q.~Liu, S.~Wu, L.~Wang, and T.~Tan, ``A dynamic recurrent model for next
  basket recommendation,'' in \emph{Proceedings of the 39th International ACM
  SIGIR conference on Research and Development in Information Retrieval}, 2016,
  pp. 729--732.

\bibitem{bai2018attribute}
T.~Bai, J.-Y. Nie, W.~X. Zhao, Y.~Zhu, P.~Du, and J.-R. Wen, ``An
  attribute-aware neural attentive model for next basket recommendation,'' in
  \emph{The 41st International ACM SIGIR Conference on Research and Development
  in Information Retrieval}, 2018, pp. 1201--1204.

\bibitem{yang2019pre}
J.~Yang, J.~Xu, J.~Tong, S.~Gao, J.~Guo, and J.~Wen, ``Pre-training of
  context-aware item representation for next basket recommendation,''
  \emph{arXiv preprint arXiv:1904.12604}, 2019.

\bibitem{fossil}
R.~He and J.~McAuley, ``Fusing similarity models with markov chains for sparse
  sequential recommendation,'' in \emph{2016 IEEE 16th International Conference
  on Data Mining (ICDM)}.\hskip 1em plus 0.5em minus 0.4em\relax IEEE, 2016,
  pp. 191--200.

\bibitem{beacon}
D.-T. Le, H.~W. Lauw, and Y.~Fang, ``Correlation-sensitive next-basket
  recommendation,'' in \emph{IJCAI International Joint Conference on Artificial
  Intelligence}, 2019.

\bibitem{adaloyal}
M.~Wan, D.~Wang, J.~Liu, P.~Bennett, and J.~McAuley, ``Representing and
  recommending shopping baskets with complementarity, compatibility and
  loyalty,'' in \emph{Proceedings of the 27th ACM International Conference on
  Information and Knowledge Management}, 2018, pp. 1133--1142.

\bibitem{Wang2021}
W.~Wang and L.~Cao, ``Interactive sequential basket recommendation by learning
  basket couplings and positive/negative feedback,'' \emph{ACM Transactions on
  Information Systems (TOIS)}, vol.~39, no.~3, pp. 1--26, 2021.

\bibitem{ugru}
T.~Donkers, B.~Loepp, and J.~Ziegler, ``Sequential user-based recurrent neural
  network recommendations,'' in \emph{Proceedings of the eleventh ACM
  conference on recommender systems}, 2017, pp. 152--160.

\bibitem{itemvec}
O.~Barkan and N.~Koenigstein, ``Item2vec: neural item embedding for
  collaborative filtering,'' in \emph{2016 IEEE 26th International Workshop on
  Machine Learning for Signal Processing (MLSP)}.\hskip 1em plus 0.5em minus
  0.4em\relax IEEE, 2016, pp. 1--6.

\bibitem{prodvec}
F.~Vasile, E.~Smirnova, and A.~Conneau, ``Meta-prod2vec: Product embeddings
  using side-information for recommendation,'' in \emph{Proceedings of the 10th
  ACM Conference on Recommender Systems}, 2016, pp. 225--232.

\bibitem{wordvec}
T.~Mikolov, I.~Sutskever, K.~Chen, G.~S. Corrado, and J.~Dean, ``Distributed
  representations of words and phrases and their compositionality,'' in
  \emph{Advances in neural information processing systems}, 2013, pp.
  3111--3119.

\bibitem{caser}
J.~Tang and K.~Wang, ``Personalized top-n sequential recommendation via
  convolutional sequence embedding,'' in \emph{Proceedings of the Eleventh ACM
  International Conference on Web Search and Data Mining}, 2018, pp. 565--573.

\bibitem{sasrec}
W.-C. Kang and J.~McAuley, ``Self-attentive sequential recommendation,'' in
  \emph{2018 IEEE International Conference on Data Mining (ICDM)}.\hskip 1em
  plus 0.5em minus 0.4em\relax IEEE, 2018, pp. 197--206.

\bibitem{bert4rec}
F.~Sun, J.~Liu, J.~Wu, C.~Pei, X.~Lin, W.~Ou, and P.~Jiang, ``Bert4rec:
  Sequential recommendation with bidirectional encoder representations from
  transformer,'' in \emph{Proceedings of the 28th ACM international conference
  on information and knowledge management}, 2019, pp. 1441--1450.

\bibitem{hgn}
C.~Ma, P.~Kang, and X.~Liu, ``Hierarchical gating networks for sequential
  recommendation,'' in \emph{Proceedings of the 25th ACM SIGKDD International
  Conference on Knowledge Discovery and Data Mining}, 2019, pp. 825--833.

\bibitem{ham}
B.~Peng, Z.~Ren, S.~Parthasarathy, and X.~Ning, ``{HAM}: Hybrid associations
  models for sequential recommendation,'' \emph{IEEE Transactions on Knowledge
  and Data Engineering}, no.~01, p. early access, jan 2021.

\bibitem{session}
S.~Wang, L.~Cao, Y.~Wang, Q.~Z. Sheng, M.~A. Orgun, and D.~Lian, ``A survey on
  session-based recommender systems,'' \emph{ACM Computing Surveys (CSUR)},
  vol.~54, no.~7, pp. 1--38, 2021.

\bibitem{gru4rec}
B.~Hidasi, A.~Karatzoglou, L.~Baltrunas, and D.~Tikk, ``Session-based
  recommendations with recurrent neural networks,'' \emph{arXiv preprint
  arXiv:1511.06939}, 2015.

\bibitem{gru4rec+}
B.~Hidasi and A.~Karatzoglou, ``Recurrent neural networks with top-k gains for
  session-based recommendations,'' in \emph{Proceedings of the 27th ACM
  International Conference on Information and Knowledge Management}, 2018, pp.
  843--852.

\bibitem{narm}
J.~Li, P.~Ren, Z.~Chen, Z.~Ren, T.~Lian, and J.~Ma, ``Neural attentive
  session-based recommendation,'' in \emph{Proceedings of the 2017 ACM on
  Conference on Information and Knowledge Management}, 2017, pp. 1419--1428.

\bibitem{stamp}
Q.~Liu, Y.~Zeng, R.~Mokhosi, and H.~Zhang, ``Stamp: short-term attention/memory
  priority model for session-based recommendation,'' in \emph{Proceedings of
  the 24th ACM SIGKDD International Conference on Knowledge Discovery and Data
  Mining}, 2018, pp. 1831--1839.

\bibitem{srgnn}
S.~Wu, Y.~Tang, Y.~Zhu, L.~Wang, X.~Xie, and T.~Tan, ``Session-based
  recommendation with graph neural networks,'' in \emph{Proceedings of the AAAI
  Conference on Artificial Intelligence}, vol.~33, no.~01, 2019, pp. 346--353.

\bibitem{fgnn}
R.~Qiu, J.~Li, Z.~Huang, and H.~Yin, ``Rethinking the item order in
  session-based recommendation with graph neural networks,'' in
  \emph{Proceedings of the 28th ACM International Conference on Information and
  Knowledge Management}, 2019, pp. 579--588.

\bibitem{pinner}
A.~Pal, C.~Eksombatchai, Y.~Zhou, B.~Zhao, C.~Rosenberg, and J.~Leskovec,
  ``Pinnersage: multi-modal user embedding framework for recommendations at
  pinterest,'' in \emph{Proceedings of the 26th ACM SIGKDD International
  Conference on Knowledge Discovery \& Data Mining}, 2020, pp. 2311--2320.

\bibitem{mcfm}
Y.~Koren, ``Factorization meets the neighborhood: a multifaceted collaborative
  filtering model,'' in \emph{Proceedings of the 14th ACM SIGKDD international
  conference on Knowledge discovery and data mining}, 2008, pp. 426--434.

\bibitem{sarwar2001item}
B.~Sarwar, G.~Karypis, J.~Konstan, and J.~Riedl, ``Item-based collaborative
  filtering recommendation algorithms,'' in \emph{Proceedings of the 10th
  international conference on World Wide Web}, 2001, pp. 285--295.

\bibitem{herd}
\BIBentryALTinterwordspacing
T.~Kameda and R.~Hastie, \emph{Herd Behavior}.\hskip 1em plus 0.5em minus
  0.4em\relax American Cancer Society, 2015, pp. 1--14. [Online]. Available:
  \url{https://onlinelibrary.wiley.com/doi/abs/10.1002/9781118900772.etrds0157}
\BIBentrySTDinterwordspacing

\bibitem{dean2004mapreduce}
J.~Dean and S.~Ghemawat, ``Mapreduce: Simplified data processing on large
  clusters,'' 2004.

\bibitem{adagrad}
J.~Duchi, E.~Hazan, and Y.~Singer, ``Adaptive subgradient methods for online
  learning and stochastic optimization.'' \emph{Journal of machine learning
  research}, vol.~12, no.~7, 2011.

\bibitem{guo2003knn}
G.~Guo, H.~Wang, D.~Bell, Y.~Bi, and K.~Greer, ``Knn model-based approach in
  classification,'' in \emph{OTM Confederated International Conferences" On the
  Move to Meaningful Internet Systems"}.\hskip 1em plus 0.5em minus 0.4em\relax
  Springer, 2003, pp. 986--996.

\bibitem{cho2011friendship}
E.~Cho, S.~A. Myers, and J.~Leskovec, ``Friendship and mobility: user movement
  in location-based social networks,'' in \emph{Proceedings of the 17th ACM
  SIGKDD international conference on Knowledge discovery and data mining},
  2011, pp. 1082--1090.

\bibitem{shan}
H.~Ying, F.~Zhuang, F.~Zhang, Y.~Liu, G.~Xu, X.~Xie, H.~Xiong, and J.~Wu,
  ``Sequential recommender system based on hierarchical attention network,'' in
  \emph{IJCAI International Joint Conference on Artificial Intelligence}, 2018.

\bibitem{coldStrat}
\BIBentryALTinterwordspacing
B.~Lika, K.~Kolomvatsos, and S.~Hadjiefthymiades, ``Facing the cold start
  problem in recommender systems,'' \emph{Expert Syst. Appl.}, vol.~41, no.~4,
  p. 2065–2073, Mar. 2014. [Online]. Available:
  \url{https://doi.org/10.1016/j.eswa.2013.09.005}
\BIBentrySTDinterwordspacing

\bibitem{jain2019attention}
S.~Jain and B.~C. Wallace, ``Attention is not explanation,'' \emph{arXiv
  preprint arXiv:1902.10186}, 2019.

\bibitem{dacrema2019we}
M.~F. Dacrema, P.~Cremonesi, and D.~Jannach, ``Are we really making much
  progress? a worrying analysis of recent neural recommendation approaches,''
  in \emph{Proceedings of the 13th ACM Conference on Recommender Systems},
  2019, pp. 101--109.

\bibitem{tsne}
L.~Van~der Maaten and G.~Hinton, ``Visualizing data using t-sne.''
  \emph{Journal of machine learning research}, vol.~9, no.~11, 2008.

\bibitem{dt4sr}
Z.~Fan, Z.~Liu, L.~Zheng, S.~Wang, and P.~S. Yu, ``Modeling sequences as
  distributions with uncertainty for sequential recommendation,'' \emph{arXiv
  preprint arXiv:2106.06165}, 2021.

\bibitem{charu2016}
C.~C. Aggarwal, \emph{Recommender Systems: The Textbook}, 1st~ed.\hskip 1em
  plus 0.5em minus 0.4em\relax Springer Publishing Company, Incorporated, 2016.

\end{thebibliography}

\vfill
\vspace*{-3\baselineskip}

%
\begin{IEEEbiographynophoto}{Bo~Peng}
%
 is a Ph.D. student at the Computer
 Science and Engineering Department, The Ohio State University.
 His research interests include machine learning, data mining and their applications in
 recommender systems and graph mining.

\end{IEEEbiographynophoto}
\vspace*{-2.5\baselineskip}
\begin{IEEEbiographynophoto}{Zhiyun~Ren}
  received her Ph.D. degree from the Department of Computer Science, 
  George Mason University, Fairfax, in 2019. 
  Her research interests include machine learning, data mining and their applications 
  in learning analytics, recommender systems and biomedical informatics. 
\end{IEEEbiographynophoto}
\vspace*{-2.5\baselineskip}
\begin{IEEEbiographynophoto}{Srinivasan~Parthasarathy}
received his Ph.D. degree from the Department of Computer Science,
University of Rochester, Rochester, in 1999. He is currently 
a Professor at the Computer Science and Engineering Department, and 
the Biomedical Informatics Department, The Ohio State University.
His research is on high performance data analytics, graph analytics and network science, and machine learning and database systems.

\end{IEEEbiographynophoto}
\vspace*{-2.5\baselineskip}
\begin{IEEEbiographynophoto}{Xia~Ning}
  received her Ph.D. degree from the Department of Computer Science \& Engineering,
  University of Minnesota, Twin Cities, in 2012. She is currently 
  an Associate Professor at the Biomedical Informatics Department, and the Computer
  Science and Engineering Department, The Ohio State University. Her
  research is on data mining, machine learning and artificial intelligence with applications 
  in recommender systems, drug discovery and medical informatics. 
\end{IEEEbiographynophoto}
\vfill

\end{document}


\title{\methodAll: Mixed Models with Preferences, Popularities and Transitions for Next-Basket Recommendation (Supplementary Materials)}


\author{Bo~Peng,
        Zhiyun~Ren,
        Srinivasan~Parthasarathy,~\IEEEmembership{Member,~IEEE,}
        and~Xia~Ning$^*$,~\IEEEmembership{Member,~IEEE}
\IEEEcompsocitemizethanks{
\IEEEcompsocthanksitem Bo Peng 
is with the Department
of Computer Science and Engineering, The Ohio State University, Columbus,
OH, 43210.\protect\\
E-mail: peng.707@buckeyemail.osu.edu 
\IEEEcompsocthanksitem Srinivasan Parthasarathy and Xia Ning are with the Department of Biomedical Informatics, the Department of Computer Science and Engineering, 
and the Translational Data Analytics Institute,
The Ohio State University, Columbus, OH, 43210.\protect\\
E-mail:  srini@cse.ohio-state.edu, ning.104@osu.edu
\IEEEcompsocthanksitem Zhiyun Ren is with the Department of Biomedical Informatics, 
The Ohio State University, Columbus, OH, 43210.\protect\\
E-mail: ren.685@osu.edu
\IEEEcompsocthanksitem $^*$Corresponding author
}
\thanks{Manuscript received April 19, 2005; revised August 26, 2015.}}

\markboth{Journal of \LaTeX\ Class Files,~Vol.~14, No.~8, August~2015}%
{Shell \MakeLowercase{\textit{et al.}}: Bare Demo of IEEEtran.cls for Computer Society Journals}

\maketitle

\setcounter{table}{0}
\renewcommand{\thetable}{S\arabic{table}}
\setcounter{section}{0}
\renewcommand{\thesection}{S\arabic{section}}

\section{Next-Basket Recommendation vs Sequential Recommendation}
\label{sec:dis:sequence}

In this paper, we focus on the problem of next-basket recommendation.
We do not consider sequential recommendation methods as baselines 
in our experiments 
due to the fact that sequential recommendation methods usually assume that single items 
are interacted/purchased at different timestamps. 
This assumption does not really hold in next-basket recommendations that items could be 
interacted/purchased at the same timestamp.
Thus, sequential recommendation methods might not be easily adapted for next-basket recommendations.
However, our method can be extended to the sequential recommendation problem by designing an appropriate objective and replacing basket-level embeddings 
(Equation~\ref{eqn:latent} in the main text) with item embeddings in each timestamp to be consistent with the setting of the sequential recommendation, 
as in the literature~{\mbox{\cite{sasrec,caser,hgn,ham}}}.
We leave the investigation of extending \methodAll to the sequential recommendation as in our future work.

\section{Performance on the order-based split protocol}
\label{sec:exp:res:order}

%
\input{performanceLastOne.tex}

We apply a widely used protocol~{\mbox{\cite{dream,fpmc,ham,caser}}} 
to split training, validation and testing sets based on the sequential order, 
and evaluate the methods on the last basket recommendation task.
Specifically, in all the datasets, for each user, we use her/his last and second last basket as the testing and validation basket, respectively.
We use the other baskets of each user for training and measure the training error on the last basket of the training data (i.e., the third last basket 
in the original sequences before split).
We tune the parameters using grid search and use the best parameters in terms of recall@5 
on the validation set during testing for the {\methodAll} and all the baseline methods.
Table~{\ref{tbl:performanceLastOne}} presents the overall performance of all the methods in recommending the last basket.
In this setting, we only consider users with at least 4 baskets.
The number of users used in each dataset is in the parentheses after the dataset in the table.
As shown in Table~{\ref{tbl:performanceLastOne}}, the performance of {\methodAll} and baseline methods in this setting has a similar trend as that 
in the time-based split settings in the main paper.
Overall, {\method} is still the best performing method.
In terms of recall@$k$, {\method} achieves the best performance on all the datasets with a significant improvement over the best baseline methods.
In terms of NDCG@$k$, {\method} could still achieve the best performance on 3 out of 4 datasets, and achieve the second best performance on the TMall dataset.
In this setting, overall, {\FREQ} achieves similar performance with {\FREQP}, and both of them substantially outperform the best baseline methods on TMall, sTMall and Gowalla datasets.
On the TaFeng dataset, {\FREQ} still achieves superior performance over the best baseline method {\SetsSets}.
These results demonstrate the effectiveness of {\methodAll} under the widely used but questionable order-based split protocol.

\section{Parameters for Reproducibility}
\label{sec:app:para}

\input{parameter_all.tex}

We implemented \methodAll in python 3.7.3 with PyTorch 1.4.0 (https://pytorch.org).
We used Adagrad optimizer with learning rate 1e-2 on all the datasets in all the tasks.
We initialized all the learnable parameters using the default initialization methods 
in PyTorch~\footnote{{\url{https://discuss.pytorch.org/t/whats-the-default 
-initialization-methods-for-layers/3157}}}.
The dimension of the hidden representation $d$, 
the time-decay parameter $\gamma$, 
and the regularization parameter $\lambda$ that are specific for each dataset 
are reported in the \method column of Table~\ref{tbl:para:all}.
During the grid search, we initially searched $d$ from $\{8, 16, 32, 64, 128\}$, 
$\gamma$ from $\{0.2, 0.4, 0.6, 0.8, 1.0\}$, 
and $\lambda$ from $\{\text{1e-5}, \text{1e-4}, \text{1e-3}, \text{1e-2}\}$ on all the datasets for all the methods which have the corresponding parameters.
After that, if a parameter yields the best performance on the validation set when its value is on the boundary of the search range, 
we will extend the search range of this parameter, if applicable, 
until a value in the middle yields the best performance, while fixing the range of the other parameters.

For \SetsSets, we used the implementation provided by the authors in 
GitHub~\footnote{\mbox{\url{https://github.com/HaojiHu/Sets2Sets}}}.
We used the default Adam optimizer with learning rate 1e-4.
We also used the default weight 10 for the partitioned set
margin constraint.
The other parameters that are dataset specific are reported in the \SetsSets column of Table~\ref{tbl:para:all}.

For \Dream and \FPMC, since we did not find available implementations provided by the authors online, 
we implemented \Dream and \FPMC by ourself.
We implemented \Dream and \FPMC in python 3 with pytorch 1.4.0 (https://pytorch.org).
We used Adagrad optimizer with learning rate 1e-2 on all the datasets in all the tasks.
The other parameters that are dataset specific are 
reported in the \Dream and \FPMC columns of Table~\ref{tbl:para:all}.
The implementations of \method, \Dream and \FPMC is available in GitHub~\footnote{\url{https://github.com/ninglab/M2}}. 

\section{Overall performance at precision@$k$}
\label{sec:exp:res:precision}

\subsection{Overall performance at precision@$k$ on the first next basket}
\label{sec:exp:res:precision:first}
\input{performanceNextOne_precision}
Table~{\ref{tbl:performanceNextOne:precision}} presents the performance of different methods at precision@$k$ in recommending the first next basket.
In this table, for each dataset, the best performance among {\methodAll} variants (i.e., {\FREQ}, {\FREQP} and {\method}) is in bold, 
the best performance among baseline methods (e.g., {\POP}, {\POEP}, {\SetsSets}) is underlined, 
and the overall best performance is indicated by a dagger (i.e., $^\dagger$). 
Overall, the trend shown in Table~{\ref{tbl:performanceNextOne:precision}} is very similar to that shown in Table~{\ref{tbl:performanceNextOne}} in the main text.
In terms of precision@$k$, {\method} is still the best performing method, which achieves the best performance on all the 4 datasets.
Compared to the best baseline method on each dataset, {\method} consistently achieves statistically significant improvement on all the datasets over all the metrics.
These results strongly demonstrate the superior performance of {\method} over the baseline methods.

\subsection{Overall performance at precision@$k$ on the second next basket}
\label{sec:exp:res:precision:second}

\input{performanceNextTwo_precision}

Table~{\ref{tbl:performanceNextTwo:precision}} presents the overall performance at precision@$k$ in recommending the second next basket.
Generally, the results in Table~{\ref{tbl:performanceNextTwo:precision}} also show very similar trend with that in Table~{\ref{tbl:performanceNextTwo}} in the main text.
{\method} is still the best performing method in this task at precision@$k$.
{\method} achieves the best performance on 3 out of 4 datasets, and achieves the second best performance on the TaFeng dataset.
{\FREQP} is the second best performing method. 
It achieves the best performance on TaFeng, and the second best or (near) the second best performance on the TMall, sTMall and Gowalla datasets.
It is also worth noting that compared to the best baseline method on each dataset, the best {\methodAll} variant (i.e., {\FREQ}, {\FREQP} or {\method}) 
achieves statistically significant improvement on all the datasets at all the metrics except the precision@$5$ in Gowalla.

\subsection{Overall performance at precision@$k$ on the third next basket}
\label{sec:exp:res:precision:third}

\input{performanceNextThree_precision}

Table~{\ref{tbl:performanceNextThree:precision}} presents the overall performance at precision@$k$ in recommending the third next basket.
Similar to the trend in recommending the first, and second next basket, 
in the task of recommending the third next basket, {\method} is still the best performing method.
In terms of precision@$5$, it achieves the best performance on 3 out of 4 datasets.
On the TaFeng dataset, it also achieves near the second best performance.
{\FREQP} is still the second best performing method, which achieves the best performance on TaFeng and sTMall, and the second best performance on the TMall and Gowalla datasets.
Compared to the best baseline methods, in this task, the best {\methodAll} variant (i.e., {\FREQ}, {\FREQP} or {\method}) still achieves statistically significant improvement over the best baseline methods at all the metrics 
on TaFeng, sTMall and Gowalla.
On the TMall dataset, {\method} also achieves statistically significant improvement over the best baseline method {\POEP} at precision@$5$.

\section{Analysis on the diversity of recommendations}
\label{sec:exp:ana:pop}

\input{pop_dis}


We also evaluate the diversity of the recommendations generated by different methods.
Specifically, for each method, we consider the top-20 recommend items for each user, 
and then bin the recommended items into different buckets based on their frequencies in the dataset.
In this analysis, we have 10 buckets in total: 
5 buckets are for the top-10\%, top 10-20\%, top 20-30\%, top 30-40\% and top 40-50\% most frequent items; 
the other 5 buckets are for the bottom 40-50\%, bottom 30-40\%, \ldots, bottom-10\% most frequent items.
%
We conduct the analysis on the widely used TaFeng and TMall datasets. 
All the methods are trained using the training and validation baskets for the task of 
recommending the first next basket.
The frequencies of items are also calculated from the training and validation baskets.
In this analysis, for the baseline methods, 
we do not consider {\POP} since {\POP} only recommends the most frequent items (Section~{\ref{sec:exp:baselines}} in the main text).
Among the {\methodAll} variants, we consider the two best performing methods {\FREQP} and {\method}.

Table~{\ref{tbl:pop}} presents the percentage distributions of recommended items in different buckets. 
For example, on TaFeng, for {\method}, 
12.3\% of the recommended items are among the top-10\% most frequent items.
%
As shown in Table~{\ref{tbl:pop}}, 
{\method} could generate more diverse recommendations compared to 
the model-based baseline methods (i.e., {\Dream}, {\FPMC} and {\SetsSets}).
For example, for {\method}, 12.3\% of the recommended items 
are among the top-10\% most frequent items, 
while for {\Dream}, {\FPMC} and {\SetsSets}, more recommender items, that is, 100.0\%, 19.4\% and 13.4\%, 
respectively, are among the top-10\% most frequent items.
That is, on TaFeng, {\method} recommends fewer most 
frequent items compared to these baseline methods. 
This result indicates better diversity among the recommended items generate by {\method}.
We also found a similar trend in TMall.
On TMall, 
10.3\% of the recommended items from {\method} are among the top-10\% most frequent items, 
while for {\Dream}, {\FPMC} and {\SetsSets}, the percentage increases to 71.1\%, 12.8\% and 10.6\%, respectively.
We noticed that the baseline method {\POEP} recommends slightly more 
diverse items compared to {\method}.
For example, on TaFeng, 10.2\% of the recommended items from {\POEP} are among the top-10\% most frequent items, 
while for {\method}, the percentage increases to 12.3\%. 
However, as shown in Table~{\ref{tbl:performanceNextOne}} (main text) and Table~{\ref{tbl:performanceNextOne:precision}},
{\method} statistically significantly outperforms {\POEP} on all the datasets over most of the metrics.
Considering both the diversity and quality of the recommendations, 
{\method} could still substantially outperform {\POEP} in real applications. 

\section{Limitations of the Dunnhumby and Instacart datasets}
\label{sec:datasets:limitation}

We notice that in the experiments of \SetsSets, the authors used 
the Dunnhumby dataset\footnote{\url{https://www.dunnhumby.com/source-files/}}. 
However, since this dataset is simulated, the results on this dataset may not necessarily represent 
the models' performance in real applications. Thus, we {do not} use this dataset in our experiments.
It is worth noting that, although we do not use Dunnhumby in our final experiments, 
we compared {\methodAll} and baseline methods on this dataset in our preliminary study, and found that 
{\methodAll} performs at least similarly to the best baseline method.
For example, in terms of recall@5 (Section~{\ref{sec:exp:metric}} in the main text), {\method} achieves the best performance at 0.1500, 
and the best baseline method {\POEP} achieves similar performance as 0.1499.

The Instacart dataset\footnote{\mbox{\url{https://www.instacart.com/datasets/grocery-shopping-2017}}} is another dataset used in the literature~\mbox{\cite{adaloyal}}.
However, 
the original version of this dataset is not publicly available now. 
We found on the Kaggle dataset~\footnote{\mbox{\url{https://www.kaggle.com/c/instacart-market-basket-analysis/data}}} that
there is no absolute time information for each basket, and we cannot conduct time-based split as that presented in the main text (Section~{\ref{sec:exp:proto}}).
Considering the above limitations, we also do not use the Instacart dataset in our experiments.

\section{Intra-Basket Item Complementarities}
\label{sec:dis:synergies}

We noticed that some recent publications~\cite{adaloyal} assume that
items in the same basket are complementary and show that modeling this pattern could slightly improve the recommendation performance. 
However, this pattern 
actually highly depends on the datasets and how the baskets are defined. 
For example, in the online shopping scenario (i.e., the most popular recommendation scenario), 
the baskets are usually defined as all the purchased items in one ``cart". 
The items in the same ``cart" could be added in very different timestamps. 
Thus, it might not always be reasonable to assume they are complementary. 
Based on our experiments and the existing literature,
we did not find concrete evidences 
to show that most of the items in a basket could be complementary.
Actually, we also tried to model the intra-basket item complementarities in \methodAll 
by regularizing items in the same baskets to have similar embeddings.
However, it did not improve the performance on benchmark datasets, 
while significantly increased the training time. 
Considering the above reasons, 
we did not model the intra-basket item complementarities in \methodAll.

\bibliographystyle{IEEEtran}
\bibliography{paper}